\documentclass{ieeeaccess}
\usepackage{cite}
\usepackage{amsmath,amssymb,amsfonts}
\usepackage{graphicx}
\usepackage{textcomp}

\usepackage{booktabs} 
\usepackage{url}
\usepackage{caption}
\renewcommand{\figurename}{FIGURE}
\usepackage{multirow}
\usepackage{tabularx}

\def\BibTeX{{\rm B\kern-.05em{\sc i\kern-.025em b}\kern-.08em
		T\kern-.1667em\lower.7ex\hbox{E}\kern-.125emX}}

\begin{document}
	\history{Received June 18, 2020, accepted June 23, 2020. Date of publication xxxx 00, 0000, date of current version xxxx 00, 0000.}
	\doi{10.1109/ACCESS.2020.3004964}
	\title{Learning Combinatorial Optimization on Graphs: A Survey with Applications to Networking}
	\author{ \uppercase{Natalia Vesselinova}\authorrefmark{1}, (Member, IEEE), 
		\uppercase{Rebecca~Steinert}\authorrefmark{1}, 
		\uppercase{Daniel F. Perez-Ramirez}\authorrefmark{1}, 
		\uppercase{Magnus Boman}\authorrefmark{2}, (Member, IEEE)
		}
	\address[1]{Research Institutes of Sweden, RISE AB, 164 40 Kista, Sweden}
	\address[2]{Royal Institute of Technology, KTH, 164 40 Kista, Sweden}
	
	\tfootnote{This work has been financially supported by the Swedish Foundation for Strategic Research (SSF) Time Critical Clouds (Grant~No.~RIT15-0075) and Celtic Plus 5G-PERFECTA (Vinnova, Grant~Agreement~No.~2018-00735).}

	\markboth
	{Vesselinova \headeretal: Learning Combinatorial Optimization on Graphs: A Survey with Applications to Networking}
	{Vesselinova \headeretal: Learning Combinatorial Optimization on Graphs: A Survey with Applications to Networking}
	
	\corresp{Corresponding author: Natalia~Vesselinova (e-mail: natalia.vesselinova@ri.se).}
	
	\begin{abstract}
		Existing approaches to solving combinatorial optimization problems on graphs suffer from the need to engineer
		each problem algorithmically, with practical problems recurring in many instances. The practical side of theoretical computer science, such as computational complexity, then needs to be addressed.
		Relevant developments in machine learning research on  graphs are surveyed for this purpose.
		We organize and compare the structures involved with learning to solve combinatorial optimization problems, with a special eye on the telecommunications domain and its continuous development of live and research networks.
	\end{abstract}
	
	\begin{keywords}
		combinatorial optimization, machine learning, deep learning, graph embeddings, graph neural networks, attention mechanisms, reinforcement learning, communication networks, resource management.		
	\end{keywords}

	\maketitle
	
	\section{Introduction}
	\label{sec:intro}
	
	\PARstart{C}{ombinatorial} optimization problems  
	arise in various and heterogeneous domains such as routing, scheduling, planning, decision-making processes, transportation and telecommunications, and therefore have a direct impact on practical scenarios\cite{festa2014brief}.  
	Existing approaches suffer from certain limitations when applied to practical problems:  
	forbidding execution time and the need to hand engineer algorithmic rules for each separate problem. 
	The latter requires substantial domain knowledge, advanced theoretical skills, and considerable development  effort and time.
	At the same time, the ability to run efficient algorithms is crucial for solving  present-day large-scale combinatorial optimization challenges encountered in many established and emerging heterogeneous areas.  
	{R}{ecent} years have seen a surge in the development of the machine learning field and especially in the deep learning and deep reinforcement learning areas. This has led to dramatic performance improvements on many tasks within diverse areas. 
	The machine learning accomplishments together with the imperative need to efficiently solve combinatorial optimization problems in practical scenarios (in terms of execution time and quality of the solutions) are a major driving force for devising innovative solutions to combinatorial challenges.  
	We note that  the inherent structure of the problems in numerous fields or the  data itself is that of a graph \cite{hamilton2017representation}. 
	In this light, it is of paramount interest to examine the potential of machine learning for addressing combinatorial optimization problems on graphs and in particular,  for overcoming the limitations of the traditional approaches.
	
	\subsection{Goal}
	With the present survey, we seek to answer a few relevant questions: Can machine learning automate the learning of heuristics for combinatorial optimization tasks to efficiently solve such challenges? What are the core machine learning methods employed for addressing these relevant for the practice problems? What is their applicability to practical domains? 
	In other words, our goal is to bring insights into how machine learning can be employed to solve combinatorial optimization problems on graphs and how to apply machine learning to similar challenges from the telecommunications field.

	\subsection{Contribution}
	To answer these questions we:
	\begin{itemize}
		\item Provide a brief introduction to combinatorial optimization (Section~\ref{sec:MotivationMLCOG}) and fundamental problems in this area (Appendix), as well as the specific motivating questions that have prompted machine learning interest in tackling combinatorial tasks  (Section~\ref{subsec:motivation}).
		
		\item Outline contemporary machine learning concepts and methods employed for solving combinatorial optimization problems on graphs (Section~\ref{sec:theory}).  
		
		\item Present a set of (supervised and reinforcement) learning approaches to the surveyed problems (Section~\ref{sec:MLforCOP}), which provides a basis for analysis and comparison.
		
		\item Introduce a new taxonomy based on problem setting---we summarize performance results for each model developed for solving a particular combinatorial optimization problem on graphs  (Section~\ref{sec:discuss}). This categorization brings new perspectives into understanding the models, their potential as well as limitations.  Furthermore, it allows for: (1) understanding the current state-of-the-art in the context of results produced by traditional, 
		non-learned methods and (2) understanding which tools are potentially (more) suitable for solving each class of problems. Such understanding can guide  \textit{researchers} towards aspects of the machine learning models that need further improvement and  \textit{practitioners} in their choice of models for solving the combinatorial problems they have at hand.
		
		\item Illustrate the applicability of the contemporary machine learning concepts to the telecommunication networks (Section~\ref{sec:COinNet}) as the networking domain provides a rich palette of combinatorial optimization problems.
	\end{itemize}

	\subsection{Related work}
	As a result of the accelerated  research in the machine learning area, novel and advanced techniques for solving combinatorial optimization problems have been developed in the past few years. Our focus is on these most recent advancements. For an overview of earlier contributions and the history of neural networks for combinatorial optimization (up to 1999),  see  the review of Smith~\cite{smith1999neural}.
	Lombardi and Milano~\cite{lombardi2018boosting} provide a thorough overview of the use of machine learning in the modeling component of any optimization process. In particular, this contemporary survey investigates the applicability of machine learning to enhance the optimization process by either learning single constraints, objective functions, or the entire optimization model. 
	The in-depth review of Bengio et al.~\cite{bengio2018machine} investigates every aspect of the interplay and envisioned synergy between the machine learning and combinatorial optimization fields and suggests perspective research directions at the intersection of these two disciplines based on identified present shortcomings and perceived future advantages. 
	The work of Mazyavkina et al.~\cite{mazyavkina2020reinforcement} has a more narrow focus as it explores reinforcement learning as a sole tool for solving combinatorial optimization problems.
	
	The scope of our  survey shares the same broad machine learning for combinatorial optimization  topic with the aforementioned works. However, we differ from prior art in a few important aspects. First, our interest is in learning to solve combinatorial optimization problems that can be formulated on graphs because many real-world problems are defined on graphs~\cite{hamilton2017representation}. In contrast, Bengio et al.~\cite{bengio2018machine} focus on any $\mathcal{NP}$-hard combinatorial optimization problem. 
	Mazyavkina et al.~\cite{mazyavkina2020reinforcement} investigate  reinforcement learning as a sole tool for approximating combinatorial optimization problems of any kind (not specifically those defined on graphs), whereas we survey all machine learning methods developed or applied for solving combinatorial optimization problems with focus on those tasks formulated on graphs. 
	We also differ in the audience, who for Bengio et al.~\cite{bengio2018machine} and Mazyavkina et al.~\cite{mazyavkina2020reinforcement} is primarily the machine learning, mathematical and operations research communities (as explicitly stated in \cite{mazyavkina2020reinforcement} and implicitly in \cite{bengio2018machine} through the specialized literature discussed therein).  We aim at bringing insights into the most recent machine learning approaches for combinatorial optimization problems in an accessible to a broad readership form (Section~\ref{sec:MotivationMLCOG} to Section~\ref{sec:discuss}), so that specialists from any field of science can benefit from them. 
	Moreover, unlike the other surveys, 
	we synthesize performance results reported in the surveyed papers. We assemble them per class of problems (Section~\ref{sec:discuss}), which fosters conditions for revealing current advantages and shortcomings of machine learning approaches when contrasted with performance results from traditional algorithms. The performance comparison between machine learning models, on the other hand, allows for discovering 
	trends and for selecting the best 
	(according to some criteria) performing model for a given problem. 
	In addition and by contrast to existing surveys, we illustrate how the machine learning structures used for solving combinatorial optimization problems on graphs can be leveraged to combinatorial problems from the networking~domain.  
	
	\section{Why learn to solve combinatorial optimization problems?}
	\label{sec:MotivationMLCOG}
	
	Formally, a combinatorial optimization problem can be defined as a set of instances $C = \{F, c\}$, where $F$ is the set of feasible solutions and $c$ is a cost function: $c: F \mapsto \mathbb{R}$. The task can be defined as: find the optimal feasible solution (\textit{optimization} version), find the  cost of the optimal solution (\textit{evaluation} version), or the task can be formulated as a question (namely as a decision problem): is there a feasible solution $f \in F$ such that $c(f) \leq L$, where $L$ is some integer (\textit{recognition} version) \cite{papadimitriou1998combinatorial}.

	The primary goal of combinatorial optimization is to devise efficient algorithms for solving such problems. In computer science, an algorithm is called \textit{efficient}  as long as the number of elementary steps of the algorithm grows as polynomial in the size of the input \cite{papadimitriou1998combinatorial}. Problems that can be solved in polynomial time by a deterministic algorithm are called problems in $\mathcal{P}$. However, most of the  
	combinatorial optimization problems are considered computationally intractable  
	since no exact polynomial-time algorithm has been devised for solving them yet. 
	A problem is in $\mathcal{NP}$ if and only if its decision problem is solvable in polynomial time by some non-deterministic algorithm. A problem is called $\mathcal{NP}$-\textit{hard} if every problem in $\mathcal{NP}$ can be reduced to it in polynomial time. A problem is called  $\mathcal{NP}$-\textit{complete} if it is in $\mathcal{NP}$ and it is $\mathcal{NP}$-hard.

	Despite that many  
	combinatorial optimization problems are $\mathcal{NP}$-hard, they arise in diverse real-life scenarios 
	spanning telecommunications, transportation, routing, scheduling, planning, and decision making 
	among many other fields, and hence have a practical impact.
	A succinct reminder of some of the most prominent and fundamental problems is provided in the Appendix. Those are also the problems we survey in this work. The selection of problems has been mostly defined by two interrelated factors: the application range of the particular combinatorial optimization task as well as whether the problem has been on the radar of the machine learning research community. 
	In the Appendix, we also provide references to some exact, approximation, or meta-heuristic algorithms that address the surveyed combinatorial tasks. For some problems such as the maximum vertex cover, lower  and upper bounds on the computational complexity have been determined (see Chen et al.~\cite{chen2006improved}, for example), whereas other problems such as the vehicular routing problem are more difficult in terms of setting such computational bounds. These studies are incorporated to set the relevant context within which combinatorial optimization problems are solved, namely examples of usual and more recent, non-learned approaches.     
	Some of the referenced works also assess the complexity of the proposed algorithms. In the introduction to Section~\ref{sec:discuss}, we provide the reader with references that can serve as guidelines for evaluating the complexity of different machine learning structures and approaches. This combined material equips the reader with the necessary sources for making further performance comparisons. 
	
	Computational complexity comes in many different forms: time-, space- and sample-complexity; worst case, average case, and canonical case. Often there is a trade-off between these forms: a lower time-bound on worst case complexity can be achieved by increasing memory power and hence space complexity, for instance. Such trade-offs make it difficult to produce practically useful baselines from theoretical investigations alone, for instance, in the form of Big~$\mathcal{O}$ expressions. In his seminal paper \cite{karp1972reducibility} from 1972, Karp discussed combinatorial optimization problems from an angle that has inspired intense research over almost 50 years. Since many combinatorial optimization problems do not have exact solutions for the general case, approximate solutions have replaced or supplemented more precise formulations. Analogously, heuristics have replaced or augmented more theoretical formulations, and this has happened often enough for meta-heuristics to evolve as an important methodological area \cite{voss2000meta}. Genetic algorithms, simulated annealing, branch and bound, dynamic programming, and several other families of algorithms are considered meta-heuristics approaches. For coming to grips with most practical combinatorial optimization problems, new possibilities for sharing and evaluating heuristics produced within the combinatorial optimization community are more important to progress than theoretical bounds. This trend is likely to continue in the future as it is almost immune to new theoretical findings on individual problems, and represents what the community thinks in that practice and theory for combinatorial optimization problems should be used in tandem for their successful implementation and innovation.
	
	A compact yet informative introduction to the subject of combinatorial optimization is provided by Festa \cite{festa2014brief}. For a thorough introduction to combinatorial optimization, see Papadimitriou and Steiglitz~\cite{papadimitriou1998combinatorial} and refer to Garey and Johnson~\cite{garey1979computers} for a book on the theory of $\mathcal{NP}$-\textit{completeness} and computational intractability.

	\subsection{Challenges faced by traditional approaches}
	Due to the relevance of this class of problems, a rich literature on the subject has been developed in the past few decades. Exact algorithms, exhausting all possible solutions by enumeration, exhibit forbidding  execution times when solving large real-life problems.
	Approximate algorithms can obtain near-optimal solutions for practical problems and in general, provide theoretical guarantees for the quality of the produced solutions. However, approximate algorithms are only of theoretical value when their time complexity is a higher-order polynomial \cite{gonzalez2007handbook}, and importantly, such approximate approaches do not exist for all real-world problems.  
	Heuristic and meta-heuristic algorithms are usually preferred in practice because they offer a balance between  execution time and solution quality. In effect, they are often (much) faster than exact solvers and approximate algorithms but they lack theoretical guarantees for the solutions they can produce. 
	Nonetheless, the design process of such heuristic methods requires specialized domain knowledge and involves trial-and-error as well as tuning. Each combinatorial optimization problem requires its own specialized algorithm. 
	Whenever a change in the problem setting occurs, the algorithm must typically be revised, and the system needs to be optimized anew. This can be impractical as most of the challenging tasks that require optimization are large-scale in practice. 
	Another relevant aspect is the increasing complexity of such problems of practical interest. Such complexity can quickly lead to prohibitive execution times, even with the fastest solvers. In practice, this can constitute a major obstacle in using them for producing optimal solutions to real-world problems with many constraints.

	\subsection{Motivation for machine learning}
	\label{subsec:motivation}
	
	In the context of the success attained by machine learning in automating the learning and consequently the solving of complex classification, prediction and decision tasks, and the presence of the enumerated challenges of existing approaches, the natural question that arises is: \textit{Can machine learning be successfully employed to learn to solve combinatorial optimization problems?}
	
	The majority of the surveyed publications aim at answering this essential question, and in what follows, we give a brief account of the specific motivation that has driven the research endeavors behind each contribution.
	Specifically, to address the aforementioned question, Vinyals et al.~\cite{PointerNets15} construct a novel model, which through supervision can learn to approximate solutions to computationally intractable combinatorial optimization problems. 
	The work of Bello et al.~\cite{NeuralCOwithRL17} is directed towards understanding how machine learning in general and deep reinforcement learning in particular can be used for addressing $\mathcal{NP}$-hard problems, specifically  the planar TSP (see Appendix).  
	Close to their research perspective is that of Kool et al.\cite{AttentionVRP19}, who do not aim to outperform state-of-the-art TSP algorithms such as Concorde~\cite{applegate2006traveling} but instead focus their effort on making progress in learning heuristics that can be applied to a broad scope of different practical problems.
	Likewise, the goal of Prates et al.~\cite{GNNforDecisionTSP19} is not to devise a specialized TSP solver but instead to investigate whether a graph neural network can learn to solve this problem with as little supervision as possible.  
	Similarly, Lemos et al.~\cite{lemos2019graph} aim to show that a simple learning structure such as a graph neural network can be trained to solve fundamental combinatorial optimization challenges such as the graph coloring problem.

	The $\mathcal{NP}$-complete propositional satisfiability (SAT) problem in computer science (see the Appendix) has a broad scope of application in areas such as combinational equivalence checking, model checking,  automatic test-pattern generation, planning and genetics \cite{marques2008practical}.
	Although Selsam et al. \cite{SATsolver19} affirm that contemporary SAT solvers have been able to solve practical tasks with variables in the order of millions, their interest remains in verifying that a neural network can be taught to solve SAT problems.

	Some machine learning researchers, such as Nazari et al.~\cite{RLforVRP18},  have even more ambitious goals. Specifically, Nazari et al. note that many exact and heuristic algorithms for VRP exist, yet producing reliable results fast is still a challenge. Therefore, in addition to automating the process of learning to solve  $\mathcal{NP}$-hard problems without any meticulously hand-crafted rules, the goal of Nazari et al. is to obtain a state-of-the-art quality of solutions and to produce these solutions within reasonable time frames~\cite{RLforVRP18}.

	Another relevant observation that drives machine learning research in combinatorial optimization is that it might not be easy, even for experts with deep domain knowledge, to detect complex patterns or specify by hand the useful properties in data as noted by Li et al.~\cite{COGNNGuidedTreeSearch18}. Therefore, Li et al. examine the potential of machine learning approaches to learn from massive real-world datasets in order to approximate solutions to $\mathcal{NP}$-hard problems. In the light of the required specialized human knowledge for the design of good heuristics, Dai et al.~\cite{COoverGraphs17} pose the question: ``Can we automate this challenging, tedious process, and learn the algorithms instead?''~\cite{COoverGraphs17},~p.1. In effect, Dai et al. develop a framework that can learn efficient algorithms for a diverse range of combinatorial optimization problems on graphs.

	The research question asked by Mittal et al. \cite{LearningHeuristicsLargeGraphs19} is whether an approximate algorithm for solving an optimization problem can be learned from a distribution of graph instances so that a problem on unseen graphs generated from the same distribution can be solved by the learned algorithm. In other words, the authors seek to understand whether learning can be automated in a way that the learned algorithm  generalizes to unseen instances from the same data generating process.

	Overall, the aforementioned contributions, which we synthesize and analyze in the sections that follow, explore and bring valuable insights into the ability of machine learning to serve as a general tool for efficiently solving combinatorial optimization problems on graphs. 
	
	\section{Specialized, contemporary machine learning methods}
	\label{sec:theory}

	We assume that the reader has a basic understanding of core machine learning principles. The fundamental concept underlying the primary building block of deep learning, namely the (artificial) neural network, is of special relevance. Different neural networks enable the learning of different data structures and representations, among which convolutional neural networks and recurrent neural networks are central for  the reminder of this survey. The books of Bishop \cite{Bishop2006}, Goodfellow et al.~\cite{Goodfellow2016}, and Murphy~\cite{Murphy2020} offer an in-depth study of machine learning.  A brief primer to the area is provided by Simone et al.~\cite{simeone2018brief}, as well as by machine learning surveys (see for instance \cite{wang2020thirty} and Fig.~1 therein).
	
	The material presented below is a brief introduction to the essence of those contemporary machine learning structures that are the basis for the learning models surveyed in Section~\ref{sec:MLforCOP}. In other words, this section is tailored for readers with a machine learning background but no familiarity with attention mechanisms, graph neural networks, and deep reinforcement learning. Readers well acquainted with these ideas and the theory behind might, without loss of continuity, proceed directly to  Section~\ref{sec:MLforCOP}.

	\subsection{Attention mechanisms}
	\label{subsec:theory-attention}
	
	Recall that a recurrent neural network (RNN) is a generalization of a feedforward network specialized for processing sequential data (such as text, audio and video as well as time series)~\cite{Goodfellow2016}. In its basic form, an RNN computes a sequence of outputs $(y_1, ..., y_n)$ from a sequence of inputs $(x_1, ..., x_n)$ by iteratively solving the equation as follows:
	\[
	h_t = f(x_t, h_{t-1}),
	\]
	or in an expanded form:
	\[
	h_t = f (W^{hx} x_t + W^{hh} h_{t-1} + b^h),
	\]
	where $h_t$ denotes a hidden unit at time step $t$, $W$ is a weight matrix shared across all hidden units and $b^h$ is the bias. The activation function $f$ is non-linear, usually the sigmoid or hyperbolic tangent (tanh) function. The main problem observed in the traditional RNNs is their short-term memory due to the vanishing gradient problem, which can occur during back propagation when the processed sequence is long.
	State-of-the-art RNNs use advanced mechanisms such as long short-term memory (LSTM) and gated recurrent units (GRUs)~\cite{cho2014learning}) to overcome this problem.

	\textit{Sequence-to-sequence} learning  \cite{sutskever2014sequence} was introduced to solve the general problem of mapping a fixed-dimensional input to a fixed-dimensional output of potentially different length when the input and output dimensions are not known a priori and can vary. The idea is to use an encoder--decoder architecture based on two LSTM (the same or different) networks, namely the first neural network maps the input sequence to a fixed-sized vector, and then the other LSTM maps the vector to the target sequence \cite{sutskever2014sequence}. An aspect of this approach is that a variable-input sequence must be compressed into a single, fixed-length vector. This can become an obstacle when the input sequences are longer than those observed during training and, in effect, can degrade the predicting performance of the model~\cite{bahdanau2014neural}. 
	
	\textit{Attention} mechanisms have emerged as a solution to the limitations of the aforementioned encoder--decoder architectures. In essence, attention allows the decoder to use any of the encoder hidden states instead of using the fixed-length vector produced by the encoder at the end of the input sequence (the last hidden state of the encoder). This idea was introduced by Bahdanau et al.~\cite{bahdanau2014neural}, who augmented the basic encoder-decoder structure by encoding the input sequence into a sequence of vectors. An additional neural network adaptively chooses (`pays attention to') a subset of the vectors (most) relevant for generating a correct output during decoding. 
	In summary, the proposed extension  allows the model to encode the input sequence to a variable length vector instead of squashing the source---regardless of its dimension---into a vector with a pre-defined length~\cite{bahdanau2014neural}. An improved predicting performance is observed as a result of the selective use of relevant information.
	
	In order to mathematically define the model, let us denote the encoder and decoder hidden states with $(e_1, ..., e_n)$ and $(d_1, ..., d_m)$, respectively. The attention vector, at any given time $i$, is computed as the affinity between the decoder state and all encoder states:
	\begin{align}
		u_j^i & = f(W_1 e_j + W_2 d_i),	\quad & j \in (1, ..., n) \label{eq:score}\\
		a^i_j & = \text{softmax}(u^i_j),		\quad & j \in (1, ..., n) \label{eq:attention}	\\
		c_i	  & = \sum_{j=1}^{n} a^i_j e_j,						  \label{eq:context}
	\end{align} 
	where $u^i_j$ scores the extent to which the input elements around position $j$ and the output at position $i$ match. Then, the attention vector $a$ is obtained by the softmax function, which normalizes the scores to sum up to 1. Lastly, the encoder states are weighted by $a$ to obtain the context vector~$c$. This vector $c$ and the decoder state $d$ are concatenated to 1)~make predictions and 2) obtain hidden states, which are the input to the recurrent model during the next step. In summary, the attention model uses the additional information provided by the context vector together with the decoder state to produce predictions, which are shown to be better than those of the sequence-to-sequence model~\cite{bahdanau2014neural}, \cite{PointerNets15}. 
	
	The model is trained (its parameters, $W_1$ and $W_2$ in (\ref{eq:score}), are learned) by maximizing the conditional probabilities $p(\mathcal{C}^\mathcal{P} | \mathcal{P}; \theta)$ of selecting the optimal solution $\mathcal{C}^\mathcal{P}$ given the input sequence $\mathcal{P} =  (P_1, ..., P_n)$ and the parameters $\theta$ (where $\theta$ accounts for $W_1$, $W_2$ as well as any other possible parameters) of the model \cite{PointerNets15}:
	\begin{equation}
		\theta^* = \arg \max_{\theta} \sum_{\mathcal{P}, \mathcal{C}^\mathcal{P}} \log p(\mathcal{C}^\mathcal{P} | \mathcal{P}; \theta).
	\end{equation}
	Once the parameters of the model have been learned $\theta^*$, they can be used to make inference: given an input sequence $\mathcal{P}$, select the output sequence with the highest probability $\hat{\mathcal{C}}^\mathcal{P} = \arg \max_{\mathcal{C^{\mathcal{P}}}} p(\mathcal{C^{\mathcal{P}}} | \mathcal{P}; \theta^*)$ \cite{PointerNets15}.

	Another novel development based on attention mechanisms is the \textit{Transformer} proposed by Vaswani et al.~\cite{attentionIsAll}. In its essence, the Transformer~\cite{attentionIsAll} is an encoder-decoder structure. However, it substantially differs from other attention frameworks in that the RNNs are replaced with a stack of self-attention layers with positional encodings, which are implemented with neural networks of fully connected layers. 
	Self-attention is a mechanism for representing an input sequence through the attention that each input element needs to pay to the other elements of that sequence. The attention function is viewed as a mapping of a query and a set of key-value pairs to an output. First, the input is represented in three different ways (query, key, and value) by multiplying it with three different (matrices of) weights. Then, the (dot) product of the query with all keys is computed, and after applying the softmax function, the weights of the values are obtained. The output is an aggregate of the weighted values. Vaswani et al.~\cite{attentionIsAll} extend the single self-attention mechanism to multi-head attention by using different linear projections of the queries, keys, and values over the same original input. The attention is applied in parallel to all of them. The resulting values are concatenated and projected once more to obtain the final values. This procedure allows the model to  consider simultaneously relevant information from different positions.
	
	The motivation behind the development of this novel yet simpler encoder-decoder architecture is three-fold: the computational complexity per layer, the potential for parallelization (the sequential nature of the RNNs is an obstacle for parallelized training of the model) and the computation for learning long-range dependences between elements in the sequence (constant for the Transformer architecture whereas for the sequence models with attention it is linear in the size of the sequence, which makes it more difficult for the latter to learn long-range dependences). The superior performance of the Transformer on translation tasks---measured in terms of run time and quality of the translation---is attributed to the aforementioned features, namely parallelization capacity and  ability to learn long-range dependencies in long sequences, respectively \cite{attentionIsAll}.

	\subsection{Graph neural networks}
	A graph $G$ in its simplest form $G = \langle V, E \rangle$ is defined by its vertices (nodes) $v \in V$ and edges (arcs) $e \in E$. Data in numerous practical applications and domains (such as communication networks, sensor networks, urban computing, computer vision, ecology, bioinformatics, neuroscience, chemoinformatics, social networks, recommender networks, (scientific) citations, and much more, see \cite{goyal2018graph}, \cite{hamilton2017representation}, \cite{lee2019attention} and references therein) can be naturally and conveniently represented as graphs. The list of practical applications with graph data and the large body of work in the deep learning and data mining communities developed for accounting for such data attest to the ubiquity of graphs as a relevant data structure. How data is represented has a direct impact on the learning and eventually, on the performance of machine learning models. Therefore, in Table~\ref{tab:graphnet}, we have collected several extensive reviews and surveys on various aspects of graph embedding techniques and graph neural networks. This subsection  is a snapshot of this major effort and its goal  is to briefly introduce the essence of those graph neural networks that we  refer to in the sections that follow.

	\begin{table*}[!ht]
		\begin{center}
			\begin{tabular}{|l|c|l|}
				\hline
				\textbf{Reference}					& \textbf{Year} & \textbf{Summary} 	  \\
				\hline
				Hamilton et al.~\cite{hamilton2017representation}& 2017	&  review of methods (including  matrix factorization-based methods, random-walk based algorithms and \\
				&				& graph neural networks) to embed individual nodes as well as entire (sub)graphs, outline of applications	 \\
				\hline
				Goyal and Ferrara~\cite{goyal2018graph}	& 	2018	& summary of graph embedding challenges, categorization of approaches into factorization methods, random walks,    \\
				&			& and deep learning, and analysis of their performance on common datasets based on open-source Python library \\
				\hline	
				Zhang et al.~\cite{zhang2018deep}	&  2018		& categorization based on model architecture:
				graph recurrent neural networks, graph convolutional networks, \\
				& &  graph autoencoders, graph reinforcement learning, and graph adversarial methods, \\
				& & analysis of differences, outline of applications\\
				\hline
				Cai et al.~\cite{cai2018comprehensive}&  2018	&  	a problem-based graph embedding taxonomy, analysis of surveyed graph embedding techniques, \\
				&			& categorization of applications enabled by graph embedding into  node, edge and graph related\\
				\hline	
				Zhang et al.~\cite{zhang2019graph}	&  2019		& survey of spectral and spatial graph convolutional networks, categorization based on applications				\\ 
				\hline
				Li et al.~\cite{lee2019attention}  	&  	2019	& introduction of three taxonomies based on problem setting (type of input and output), \\
				& &type of attention mechanism used, and the task (graph classification, link prediction), outline of challenges 		\\
				\hline
				Wu et al.~\cite{GNNSurvey}& 2020 &  categorization of graph neural networks into:  recurrent  graph  neural networks,  \\
				& & convolutional  graph  neural  networks,  graph  autoencoders, spatial-temporal graph neural networks, \\
				& & discussion of applications, summary of the open source codes, benchmark of datasets,  evaluation of models\\
				\hline
			\end{tabular}
		\end{center}
		\caption{Surveys and reviews of graph representation learning: graph embeddings and graph neural networks in their different flavors.}
		\label{tab:graphnet}
	\end{table*}

	The central assumption about graph structured data is that there exist meaningful relations between the elements of the graph, which if known can bring insights into the data and can be used for other downstream machine learning tasks (such as prediction and classification). Naive implementation of traditional feedforward, recurrent or convolutional neural networks may make simplifying  assumptions to accommodate the graph structured data in their frameworks. \textit{Graph neural networks} (GNNs)  have been introduced to overcome such limitations: they process graph input and learn the potentially complex relations as well as the rules that guide these relations. 
	The essential idea of the GNN proposed by Scarselli et al.~\cite{scarselli2008graph} and all GNNs that have been subsequently developed is to efficiently capture the (often complex) interaction between individual nodes by updating the states of the nodes. A node's hidden state is recurrently updated in \cite{scarselli2008graph} by exchanging information (node embeddings) with neighboring nodes until a stable equilibrium is reached:
	\begin{equation}
		h_v^{(t)} = \sum_{u \in \mathcal{N}(v)} f( x_v, x_{(v,u)}^e, x_u, h_u^{(t-1)}),
		\label{eq:GNNup}
	\end{equation} 
	where  the node embeddings are initialized randomly and $f()$ is an arbitrary differentiable function that is a contraction mapping\footnote{Recall that a contraction shrinks (contracts) the distance between two points.}, so that the node embeddings converge. The node hidden states are sent to a read-out layer once a convergence is reached. Nowadays, GNNs exist in different forms (see \cite{GNNSurvey}) and GNN is used as a general term to denominate neural networks that process graph structure data. 
	
	Li et al.~\cite{li2015gated} extend and modify the GNN framework by employing GRUs and back propagation through time, which removes the need to recurrently solve (\ref{eq:GNNup}) until convergence. Some advantages of this framework are: node embeddings can be initialized with node features and intermediate outputs (in the form of subgraphs)  can be used \cite{hamilton2017representation}. The update equation takes the form:
	\begin{equation*}
		h_v^{(t)} = GRU(h_v^{(t-1)}, \sum_{u \in \mathcal{N}(v)}W h_u^{(t-1)}),
		\label{eq:GNN}
	\end{equation*}
	where $W$ is a trainable weight matrix.
	
	\textit{Graph convolutional networks} (GCN) generalize the  convolution operation  to  graph data. They generate a representation of each node by aggregating the features of the node with those of its neighbors. In contrast to recurrent GNNs (an instance of which is \cite{scarselli2008graph}), which use the same graph recurrent layer and contractive constraints to update nodes representations, GCNs use a stack of convolutional layers each with its own weights. GNNs are split into two categories: spectral and spatial based \cite{GNNSurvey}. The former have their roots in graph signal processing and use filters for defining the convolutions (interpreted as removing noise from data). The latter inherit the information propagation idea of recurrent graph networks to define the convolution.  
	
	\textit{Message  passing  neural  network}  (MPNN)~\cite{gilmer2017neural}  is a  general  framework  of  spatial-based GCNs.  The graph  convolutions  are performed as  a  message  passing  process  in  which information  is interchanged between nodes through the edges that connect them. The  message  passing function is given by \cite{GNNSurvey}:
	\begin{equation*}
		h_v^{(k)} = u_k(h_v^{(k-1)}, \sum_{u \in \mathcal{N}(v)} m_k(h_v^{(k-1)}, h_u^{(k-1)}, x_{vu}^e)),
		\label{eq:GNN}
	\end{equation*}
	where $k$ is the layer index, $u$ denotes the update function and $m$--the message passing function. The hidden representation of the nodes can be passed to an output layer or the representations can be forwarded to a read-out function to produce a useful representation of the entire graph.

	Most recently, attention mechanisms have been incorporated into GNNs to improve the graph deep learning methods by allowing the model to focus on the most relevant task-related information for making decisions. The \textit{graph attention network} (GAN) proposed by Veli{\v{c}}kovi{\'c} et al.~\cite{velivckovic2017graph} is based on (stacking) a graph attention layer. The input to this layer are the node features and the produced output is another set of node features of a higher-level. These are computed through attention coefficients, which indicate the importance of the features of node $v$ to node $u$. This attention mechanism prioritizes task relevant information by aggregating neighbor node embeddings (`messages'). The aggregation is produced by defining a probability distribution over them. In its most general form, the model can drop all structural information by allowing each node to attend to every other node in the graph. Similarly to Vaswani et al.~\cite{attentionIsAll}, the authors have found that multi-head attention can stabilize the learning process and thus be beneficial. Veli{\v{c}}kovi{\'c} et al.~\cite{velivckovic2017graph} list several advantages of their approach among which are increased model capacity (due to the implicit assignment of different importance scores to nodes from the same neighborhood), increased computational efficiency (as the operation of the attention layer can be parallelized across all edges), increased interpretability and that the model can be used for inductive learning involving tasks for which it is evaluated on unseen graph instances. For other approaches that incorporate attention into the GNN, see \cite{lee2019attention}.
	
	\textit{Graph embedding} techniques aim at representing a network of nodes as low-dimensional vectors while preserving the graph structure and  node  content  information. Such information-preserving embeddings are aimed at easing the  subsequent  graph analytics  tasks  (such  as  classification,  clustering,  and  recommendation)~\cite{GNNSurvey}. Deep learning methods address a learning task from end to end, whereas graph embedding techniques first reduce the graph data into low-dimensional space and then forward the new representation to machine learning methods for downstreaming tasks, see Hamilton et al.~\cite{hamilton2017representation} and Cai et al.~\cite{cai2018comprehensive} for comprehensive overviews of graph embedding techniques.

	\subsection{(Deep) reinforcement learning}

	Reinforcement learning is goal-oriented learning from interaction and therefore conceptually different from two other primary and popular machine learning approaches: supervised and unsupervised. In contrast to them, reinforcement learning learns from interacting with an (uncertain) environment (instead of being instructed by a teacher / labeled dataset as in supervised learning) to maximize a reward function (instead of finding hidden patterns as in unsupervised learning). In addition, reinforcement learning implements exploration and exploitation mechanisms, which are not present in the aforementioned approaches\footnote{For an in-depth treatment, see the book of Sutton and Barton~\cite{SuttonBarto2018}.}. 
	
	Reinforcement learning has four basic elements: a \textit{policy} (strategy followed by the learning agent), a \textit{reward signal} (a single value, called \textit{reward}, received by the learning agent at every time step), a \textit{value function} (which could be considered as the long-term reward), and, optionally, a \textit{model} of the environment (based on which \textit{model-based} and \textit{model-free} reinforcement learning methods are differentiated). The main goal of the reinforcement learning agent  is to maximize the total reward (or \textit{return}, the expected sum of future rewards).
	
	There are two main families of approaches in reinforcement learning: tabular solution methods and approximate solution methods \cite{SuttonBarto2018}. From the rich literature on reinforcement learning methods, we succinctly explain the essence of those used in the machine learning for combinatorial optimization approaches, described in Section~\ref{sec:MLforCOP}.
	
	\textbf{Tabular methods.}
	When the state space and action set, introduced below, are small enough, the approximate value functions can be represented as tables. The methods then can often find the exact optimal solution and exact optimal policy. 
	The Markov decision process, which is a fundamental mathematical model $M$ for analytically representing the interaction between a system (such as a reinforcement learning agent) and its environment $M = \langle S, A, T, R, \gamma \rangle$, is defined by:
	\begin{itemize}
		\item a state space $S$,
		\item a set of actions $A$,
		\item a transition model $T: S \times A \rightarrow S$, $s_{t+1} = T(s_t, a_t)$,
		\item a reward function $R: S \times A \rightarrow \mathbb{R}$ , $R_{t+1} = R(s_t, a_t)$, 
		\item a discount factor (discount rate) $0 \leq \gamma \leq 1$.
	\end{itemize}
	The return is defined by:
	\begin{equation}
		G_t = \sum_{i=0}^{\infty} \gamma^i R(s_{t+i+1}, a_{t+i+1}),
		\label{eq:RL-return}
	\end{equation}
	and the goal can be formulated as:
	\begin{equation*}
		\max_\pi G_t \quad \text{such that} \quad s_{t+1} = T(s_t, \pi(s)).
	\end{equation*}
	The discount factor is a user-defined value that balances the weight given to the immediate reward with that of future rewards \cite{SuttonBarto2018}. If $\gamma = 0$, at time step $t$, the agent is interested in maximizing the immediate reward $R(s_{t+1}, a_{t+1})$ \cite{SuttonBarto2018} only. If $\gamma < 1$, the sum (\ref{eq:RL-return}) is finite, assuming the reward sequence is bounded. When $\gamma$ approaches its maximum, the agent considers future returns more strongly. In short, the discount factor $\gamma$ defines the value of the future rewards: at $i$ time steps in the future, the reward is worth $\gamma^{-i}$ times less than when it is received immediately. For $\gamma=1$, all (immediate and future) rewards are given equal weight.
	
	A \textit{policy}, $\pi$, is a mapping from the space of states to the set of actions, $\pi: S \rightarrow A$.
	A \textit{value function} $v_{\pi}: S \rightarrow \mathbb{R}$ assigns to each state $s \in S$ a single value $v_{\pi} \in \mathbb{R}$, which is a measure of the usefulness of being in state $s$ when following policy $\pi$. It is calculated as the expected return when starting in $s$ and following $\pi$ thereafter:
	\begin{equation*}
		v_{\pi}(s_t) = \mathbb{E} \Big[ \sum_{i=0}^{\infty} \gamma^i R(s_{t+i}, \pi(s_{t+i})) | s_t = s \Big].
	\end{equation*}
	
	Similarly, we can define the \textit{action-value function} $q_{\pi}$ that takes into account the impact of taking an action $a$ when being in a state $s$ and following policy $\pi$: 
	\begin{equation*}
		\begin{split}
			q_{\pi}(s_t, a_t) = & R(s_t, a_t) + \\
			& \mathbb{E} \Big[ \sum_{i=0}^{\infty} \gamma^i R(s_{t+i}, \pi(s_{t+i})) | s_t = s, a_t = a \Big].
		\end{split}
	\end{equation*}
	A value-function always obeys the recursive relation:
	\begin{equation}
		v_{\pi}(s) = R(s, \pi(s)) + \gamma v_{\pi}(T(s, \pi(s)))
	\end{equation}
	known as the Bellman equation \cite{SuttonBarto2018}. 
	
	The value function of an \textit{optimal policy} $\pi^*$ is the maximum over all possible policies:
	\begin{equation}
		v_*(s) := \max_\pi v_{\pi}(s),
		\label{eq:optimalValue-function}
	\end{equation}
	and is called the \textit{optimal value function}. Likewise, the \textit{optimal action-value function} is given by:
	\begin{equation}
		q_*(s,a) := \max_\pi q_{\pi}(s,a).
		\label{eq:optimalQ-function}
	\end{equation}
	The latter two are related through:
	\begin{equation*}
		v_*(s) = \max_{a' \in A} q_*(s,a).
	\end{equation*}
	The recursive relation for the optimal value function
	\begin{equation*}
		v_*(s) = \max_{a' \in A} [ R(s,a) + \gamma v_*(T(s,a))]
	\end{equation*}
	is known as the \textit{Bellman optimality equation}.
	
	It has been shown that a solution to the Bellman equation, when the transition function is unknown, can be found by an iterative process. Watkins devised the Q-learning algorithm~\cite{watkins1989learning} based on this fact. 
	It is a simple yet powerful method for estimating $q_*$ (\ref{eq:optimalQ-function}).
	The Q-learning algorithm involves creating a Q-table consisting of all possible combinations of states and actions. The agent updates the entries of the table according to the reward it receives when taking an action (i.e., interacting with the environment). The values in the Q-table reflect the cumulative reward assuming that the same policy will be followed thereafter. However, in real-world scenarios the Q-table can become very large and hence infeasible to construct. To overcome such a challenge, Minh et al.~\cite{Atari2013} have introduced an advanced algorithm based on Q-learning and deep neural networks, called Deep Q-Network (DQN) (recognized as a milestone in the development of the deep reinforcement learning). The DQN, which is a convolutional neural network, learns the optimal policy using end-to-end reinforcement learning. Minh et al. have also introduced several techniques to address common reinforcement learning problems such as divergence and instability. A solution often used in the works reviewed in the next section, is the experience replay buffer, which stores past sequential experiences. The buffer is randomly sampled during training to avoid temporal correlations.

	\textbf{Approximate solution methods.} In arbitrary large state spaces, it is not practical and often not feasible even under the assumption of infinite time and data, to find an optimal policy. Therefore, an approximate solution is preferred instead. REINFORCE~\cite{williams1992simple} is an approximate, policy-gradient method that learns a parametrized policy based on a gradient of some scalar performance measure $J(\theta)$ with respect to $\theta$, the parameter vector of the policy $\pi(\theta)$. 
	If the objective of the reinforcement learning can be formulated as finding the optimal parameters $\theta^*$ of the parametrized policy:
	\begin{equation*}
		\theta^* = \arg \max_{\theta}J(\theta), 
	\end{equation*}
	then $J(\theta)$ can be defined as the expectation of the return (total reward when starting at state $s$ and follow policy $\pi$). Its gradient can be calculated and used to directly improve the policy. In particular, the REINFORCE  algorithm consists of three iterative steps: 1) run the policy $\pi$, 2) calculate the gradient of the optimization objective $\Delta_{\theta}J(\theta)$, and 3) adjust the values of the parameters accordingly $\theta \leftarrow \theta + \alpha \Delta_{\theta} J(\theta)$.

	\textbf{Actor-critic methods.} Actor-critic methods are hybrid approaches that amalgamate the benefits  of value-based and policy-based methods. Value-based methods (such as DQN) are reinforcement learning algorithms that evaluate the optimal cumulative reward and aim at finding an optimal policy $\pi^*$ by obtaining an optimal value function (\ref{eq:optimalValue-function}) or optimal action-value function (\ref{eq:optimalQ-function}). Policy-based methods (such as REINFORCE) aim at estimating the optimal strategy directly by optimizing a parametric function (typically a neural network) representing the policy (the value is secondary, if calculated at all).   In actor-critic methods, the policy structure responsible for selecting the actions is known as an \textit{actor}, whereas the estimated value function, which `criticizes' the actions of the actor is known as a \textit{critic}. After the agent selects an action, the critic evaluates the new state and determines the quality of the outcome of the action. Both actor and critic rely on gradients to learn. 
	
	Asynchronous  advantage  actor  critic  (A3C) \cite{mnih2016asynchronous} employs asynchronous gradient descent for optimizing a deep neural network. DQN and other deep reinforcement algorithms that use experience replay buffers require a large amount of memory to store experience samples among other factors. The agents in A3C asynchronously act on multiple parallel instances of the environment thus avoiding the need of the experience replay buffer. This reduces correlation of the experiences and the parallel learning actors have a stabilizing effect on the training process. In addition to improved performance, the training time is reduced significantly. The synchronous version of this model, advantage actor critic (A2C), waits for each learning actor to finish its experience before conducting an update. The performance of the asynchronous and synchronous methods are comparable.

	\begin{figure*}[!htp]
		\centering
		\includegraphics[width=\linewidth]{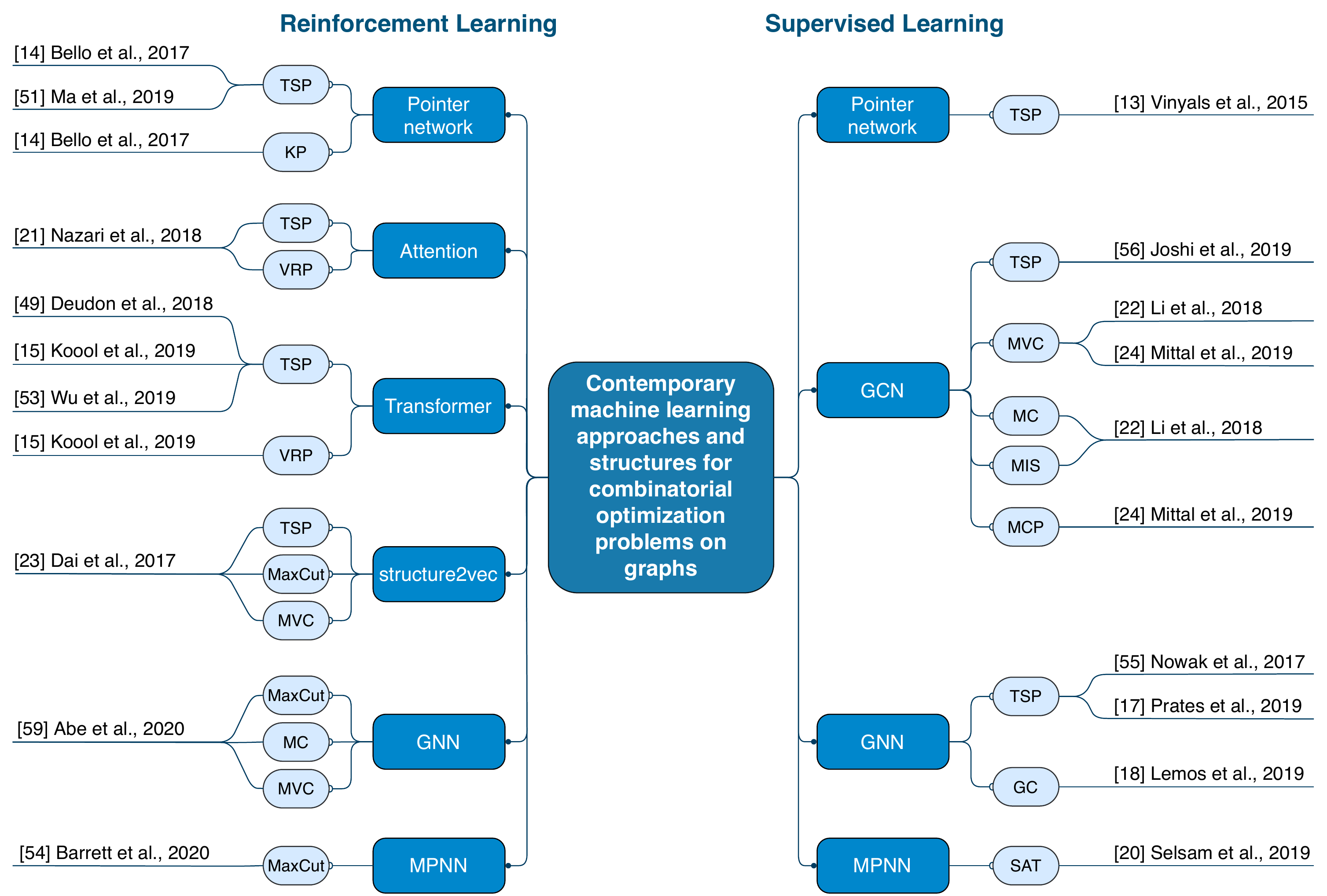}
		\caption{An overview of the contemporary machine learning approaches and structures employed to solve the surveyed combinatorial optimization problems on graphs. The categorization is based first on the type of learning---reinforcement or  supervised---followed by the type of learning structure. The combinatorial optimization problems addressed by each combination of a learning approach and a learning structure are listed together with the corresponding research contributions and their timeline.}
		\label{fig:overview} 
	\end{figure*}

	\section{Learning to solve combinatorial optimization problems on graphs}
	\label{sec:MLforCOP}

	The categorization of the machine learning models for solving combinatorial optimization problems on graphs presented below is based first on the learning structure: attention mechanisms, GNN, and their variants. Then, within each category, we differentiate the contributions based on the machine learning approach---supervised or reinforcement---and, wherever possible, a chronological order is followed.
	
	Fig.~\ref{fig:overview} depicts this categorization along with the surveyed problems, the contributions, and   the timespan of the contemporary machine learning research for combinatorial optimization up to the writing of the survey. 
	
	\subsection{Attention mechanisms: Pointer Networks and Transformer architecture} 
	\label{subsec:pointernets}
	
	\subsubsection{Supervised learning}
	The sequence-to-sequence and attention models summarized in Section~\ref{subsec:theory-attention} address some of the challenges discussed earlier: 
	the need to know a priori the dimensions of the sequences (solved by the former architecture) and the requirement to map sequences of different dimensions to a fixed-length vector (solved by the latter framework). Despite the progress made in extending the range of problems that can be tackled and in achieving improved performance as demonstrated in~\cite{sutskever2014sequence} and \cite{bahdanau2014neural}, the sequence-to-sequence model with input attention still has one limitation left---it requires the length of the output sequence to be fixed a priori~\cite{PointerNets15}. Therefore, this framework cannot be applied to the class of problems with a variable output that depends on the length of the input. Several combinatorial optimization tasks belong to this class of problems and this observation has motivated Vinyals et al.~\cite{PointerNets15} to develop a novel neural architecture called pointer network. It has proven to be a machine learning breakthrough that has served as a basis for solving diverse tasks. We summarize it below from the perspective of solving combinatorial optimization problems.
	
	The pointer network \cite{PointerNets15}  targets scenarios with discrete outputs that correspond to positions in the input. It modifies (reduces) the neural attention mechanism 
	of Bahdanau~\cite{bahdanau2014neural} as follows. Instead of blending the encoder hidden states $e_j$ into a context vector $c$ (\ref{eq:context}) at each decoder step, the proposed model~\cite{PointerNets15} uses attention to point to a member of the input sequence to be selected as the output: 
	\begin{align}
		u_j^i & = f(W_1 e_j + W_2 d_i),	\quad & j \in (1, ..., n) \\
		a	  & = \text{softmax}(u^i),			\label{eq:pointer}
	\end{align}
	where softmax normalizes the vector $u^i$ to obtain a probability distribution $a$ over the sequence of inputs~\cite{PointerNets15}. Then, $u^i_j$ are the pointers to the input elements.

	Vinyals et al.~\cite{PointerNets15} solve three non-trivial combinatorial optimization problems of geometric nature---finding planar convex hulls\footnote{The convex hull of a geometrical object  is the smallest convex set that contains the object.}, computing Delaunay triangulations\footnote{A Delanuay triangulation of a given set of discrete points $P$ in a plane is a triangulation for which points belonging to the set do not reside in the circle around any triangle in the Delaunay triangulation of $P$.}, and solving the planar (2D Euclidean) TSP---with the pointer network trained with labeled data. The authors report performance improvement over the sequence-to-sequence and attention architectures (the same architecture was used in \cite{PointerNets15} for solving the three combinatorial optimization problems without hyper-parameter tuning; in principle, such tuning might bring additional performance gains). We discuss the pointer network when applied to the planar TSP below. 
	
	An LSTM encoder is fed with a sequence of vectors, which represent the nodes that need to be visited. It generates new encodings (a new representation of each node). Another LSTM, using the pointer mechanism described earlier, produces a probability distribution $a$, see (\ref{eq:pointer}), over the nodes. The node to be visited next is the one with the highest probability. The procedure is iteratively repeated to obtain the final solution, namely a permutation over the input sequence of nodes (a tour).

	\subsubsection{Reinforcement learning}
	
	The approach proposed by Vinyals et al.~\cite{PointerNets15} was designed to tackle combinatorial optimization problems for which the output depends on the length of the input. The main limitation of \cite{PointerNets15} is that it relies on the availability of training examples as noted by Bello et al.~\cite{NeuralCOwithRL17}. First, it might be infeasible or computationally expensive to obtain labels for (large) combinatorial optimization problem instances. Second, the model performance is determined (and often limited) by the quality of the solutions (labels).
	Lastly, the supervised  approach of Vinyals et al.~\cite{PointerNets15} can only find solutions that already exist or can be generated (by the supervisor). To overcome these constraints, Bello et al.~\cite{NeuralCOwithRL17} propose to learn from experience.

	Bello et al.~\cite{NeuralCOwithRL17} tackle the TSP by using a pointer network in a fashion similar to Vinyals et al.~\cite{PointerNets15}, namely for sequentially predicting the next node from a tour. However, instead of training the model with labeled data, the authors optimize the parameters of the pointer network with model-free policy-based reinforcement learning, where the reward signal is the expected tour length. Specifically, the authors use an actor-critic algorithm that combines two different policy gradient approaches. Reinforcement learning pre-training uses the expected reward as objective. Active search does not use pre-training but begins with a random policy, optimizes the parameters of the pointer network iteratively, and retains the best solution found during search. In the former, a greedy approach is applied---during decoding the next city of the tour is the node with the highest probability. In the latter case, two different possibilities are explored: sampling, where multiple candidate tours from the stochastic policy are sampled and the shortest one is selected, and active search with or without pre-training. The parameters of the stochastic policy are refined during inference in order to minimize the loss (for active search). The advantage of active search without pre-training when contrasted to reinforcement learning with pre-training is that the former is distribution-independent, whereas the generalization capacity  of the latter depends on the training data distribution.
	
	The main constraint of the model developed by Bello et al. is that it is applicable to static problems (specifically, to the TSP in its basic form as defined in the Appendix), but not to dynamic systems that change over time (such as VRP with demands, where the demands are dynamic entities as once satisfied they become zero). To overcome this limitation, Nazari et al.~\cite{RLforVRP18} enhance the model \cite{NeuralCOwithRL17} so that it can address both static and dynamic problems (the authors focus on the VRP family).
	Specifically, Nazari et al. ~\cite{RLforVRP18} omit the encoder as neither TSP nor VRP has a naturally ordered, sequential input (any arbitrary permutation on the input nodes contains exactly the same information as the original list of nodes). Instead of an encoder, the authors embed each node by using its coordinates and demand value (a tuple of node's features) into a high-dimensional vector. Similar to the previous two approaches, Nazari et al. use an RNN decoder coupled with a particular attention mechanism. The decoder is fed with the static elements, whereas the attention layer takes as input the dynamic elements too. The variable-length alignment vector extracts from the input elements the relevant information to be used in the next decoding step. As a result, when the system state changes, the updated embeddings can be effectively calculated. Similar to  \cite{NeuralCOwithRL17}, for training the model Nazari et al. use a policy gradient approach that consists of an actor network for predicting the probability distribution of the next action and a critic network for estimating the reward. The most appealing advantage of this framework is that the learning procedure is easy to implement as long as the cost of a given solution can be computed (as it provides the reward that drives the learning of the policy). 
	
	Deudon et al.~\cite{deudon2018learning} propose a data-driven hybrid heuristic for solving TSP. They build their model upon the framework proposed by Bello et al.~\cite{NeuralCOwithRL17} by substituting the recurrent neural network with attention (used as encoder), by a Transformer architecture \cite{attentionIsAll} (recall that the latter is based solely on (multi-head) attention mechanisms, see Section~\ref{subsec:theory-attention}). The framework is further enhanced by combining the REINFORCE \cite{williams1992simple} learning rule with a 2-opt heuristic procedure \cite{croes1958method}. Deudon et al. show that by combining learned and traditional heuristics they can obtain results closer to optimality than with the model of Bello et al.~\cite{NeuralCOwithRL17}.
	
	Kool et al.~\cite{AttentionVRP19} apply the general concept of a deep neural network with attention mechanism, whose parameters are learned from experience, to solve TSP as well as VRP and their variants. In contrast to the previous three models \cite{PointerNets15}, \cite{NeuralCOwithRL17}, \cite{RLforVRP18}, which use RNNs (usually LSTMs) in the encoder-decoder architecture, the authors apply the concept of a GAN~\cite{velivckovic2017graph} (see Section~\ref{subsec: GNN}), which introduces invariance to the input order of the nodes as well as improved learning efficiency. As in the model proposed by Deudon et al.~\cite{deudon2018learning}, both the encoder and decoder are attention-based (the model is inspired by the Transformer architecture of Vaswani et al.~\cite{attentionIsAll}; the main differences between the Kool et al.~\cite{AttentionVRP19} and Deudon et al.~\cite{deudon2018learning} models are described in Appendix B in \cite{AttentionVRP19}). In fact, the attention mechanism is interpreted as a weighted message passing algorithm with which nodes interchange and extract needed information.  The input to the decoder are both the graph embedding and the nodes embeddings produced by the encoder. In addition, during the decoding step, the graph is augmented with a special context node, which consists of the graph embedding and the first and the last (from the currently constructed partial tour) output nodes. The attention layer is computed using messages only to the context node. 
	The model is trained with REINFORCE~\cite{williams1992simple} gradient estimator. At test time, two different approaches are used: \textit{greedy} decoding (which at each step takes the best, according to the model, action) or \textit{sampling}, where several solutions are sampled and the best one is reported. The results reported in \cite{AttentionVRP19} prove that more sampling improves the quality of the results but at an increased computational cost. One of the advantages of the algorithm in comparison with the RNN approaches is that the GAN enables parallelization, which explains its increased efficiency (shorter execution time, see Section~\ref{subsec:TSP}).

	The focus of Ma et al.~\cite{ma2019combinatorial} is on  large-scale TSP and time-constrained TSP. The methodology applied is similar to that of the aforementioned studies \cite{NeuralCOwithRL17}, \cite{RLforVRP18}, \cite{AttentionVRP19}, and \cite{deudon2018learning}: an encoder-decoder architecture with attention mechanism to sequentially generate a solution to the combinatorial optimization problem and learn from experience to train the model. In comparison with Bello et al.~\cite{NeuralCOwithRL17}, Ma et al. extend the pointer network with graph embedding layers and call the resulting architecture graph pointer network. Specifically, the embedding consists of (point) encoder for each city and graph embedding for the entire graph (all cities together). The latter is obtained with a GNN. The authors also add a vector context to the network with the aim to generalize to larger instances (see Section~\ref{subsec:TSP}). The vector context is applied only to large instances and consists of the vectors pointing from the current city to all other cities.
	For TSP with time constraints, the authors employ a hierarchical reinforcement learning framework inspired by Haarnoja~\cite{haarnoja2018latent}. The hierarchy consists of several (in \cite{ma2019combinatorial}, two) layers, each with its own policy and hand-engineered reward. The lower layers reward functions are designed to ensure that the solutions are in the feasible set of the constrained optimization problem, whereas the highest layer reward function adheres to the ultimate optimization objective. A  policy gradient method based on REINFORCE~\cite{williams1992simple} learns a hierarchical policy at each layer.

	Previous approaches \cite{PointerNets15}, \cite{NeuralCOwithRL17}, \cite{AttentionVRP19}, \cite{deudon2018learning}, and \cite{COoverGraphs17} (see Section~\ref{subsubsec:GNN-DRL} for an overview of \cite{COoverGraphs17}) incrementally create a solution by adding one node at each step. Wu et al.~\cite{wu2019learning} argue that learning such construction heuristics can be suboptimal since procedures such as those proposed by Kool et al.~\cite{AttentionVRP19} rely on sampling to generate multiple solutions and select the best one. However, these are generated by the same constructive heuristics. Therefore, the quality of the solution might not be improved further \cite{wu2019learning}. By contrast, Wu et al.~\cite{wu2019learning} propose a method for directly learning improvement heuristics for TSP. Such heuristics need an initial solution, which is replaced by a new one from its neighborhood in the direction of better quality in terms of optimality. This process is repeated iteratively. Usually, the  new solution is obtained by manually engineered heuristics. Wu et al.~\cite{wu2019learning} instead exploit deep reinforcement learning to obtain better improvement heuristics. The deep neural network is founded on the Transformer~\cite{attentionIsAll} and an actor-critic algorithm based on REINFORCE~\cite{williams1992simple} is employed for training. The authors note that their architecture can adopt several pairwise local operators such as the 2-opt \cite{croes1958method} heuristic.

	\subsection{Graph Neural Networks} 
	\label{subsec: GNN}
	
	\subsubsection{Supervised learning}
	
	A core element of the supervised framework proposed by Li et al. \cite{COGNNGuidedTreeSearch18} for solving $\mathcal{NP}$-hard problems, namely SAT, MIS, MVC, MC, is a GCN. The GCN is trained to estimate the likelihood of a node participating in the sought optimal solution. Since such an approach can produce more than one optimal solution and each node can participate in several solutions, the authors use a specialized structure and loss that allows them to differentiate between various solutions. The trained GCN then guides a tree search procedure, which runs in parallel. The resulting framework produces a large number of potential solutions, which are refined one at a time. The final output is the best (among all obtained) result.

	Mittal et al. \cite{LearningHeuristicsLargeGraphs19} learn to solve the influence maximization\footnote{The influence maximization problem is typical for applications within the social sciences such as viral marketing. In this context, the task is to find $k$ nodes from a graph $G$ with diffusion probabilities (represented by edge weights), that can initially receive information to maximize the influence of this information to the network (i.e., $G$).} (IM), MVC and MCP problems on billion-size graphs. The framework builds upon the architecture proposed by Dai et al.~\cite{COoverGraphs17}. It is end-to-end too, but unlike the S2V-DQN architecture \cite{COoverGraphs17}, the framework of Mittal et al. \cite{LearningHeuristicsLargeGraphs19} is supervised, which according to the authors yields higher quality predictions. It consists of two training phases: a supervised GCN that learns useful individual node embeddings (namely, embeddings that encode the effect of a node on the solution set) and a deep neural network that predicts the nodes that collectively form an optimal or close to optimal solution set. In other words, the GCN identifies the potential solution nodes and passes them to a deep neural network that learns a Q-function for predicting the solution set.

	Common for the formalism introduced by Dai et al.~\cite{COoverGraphs17} and followed by Mittal et al.~\cite{LearningHeuristicsLargeGraphs19} is that a solution is gradually built by incorporating to the solution subset one node at a time. According to Barret et al.~\cite{barrett2019exploratory} such straightforward application of Q-learning to combinatorial optimization can be suboptimal since it is challenging to learn a single function approximation of (\ref{eq:optimalQ-function}) that generalizes across all possible graphs. Therefore,  Barret et al.~\cite{barrett2019exploratory} propose an alternative exploration approach in which the reinforcement learning agent is trained to explore the solution space at test time. The Q-value (learned by a message passing neural network) of adding or removing a node from the solution set is re-evaluated and can be reversed. In short, instead of learning to construct a single solution, the agent can revise its earlier decisions and can continuously seek to improve its decisions by exploring at test time. Central for improved performance (over sequentially building a solution) is how the reward is shaped. Furthermore, Barret et al. build heuristics based on observations for deciding on the value of a node (inclusion/exclusion from the solution set). The exploratory combinatorial  optimization   approach addresses the MaxCut problem but the authors suggest that it is general enough to be applicable to any combinatorial optimization task on a graph.

	Nowak et al. \cite{nowak2017note} solve the quadratic assignment problem (TSP is an instance of it). The problem is defined by two sets (facilities and locations) of equal size $n$. A distance and a flow are defined for each pair of facilities and these two attributes define the cost. The objective is to assign each facility to a different location such that the total cost is minimized. The authors employ a GNN since the quadratic assignment problem naturally lends itself to being formulated on graphs (recall that GNNs have been specifically developed for graph structured data). Nowak et al. draw two possible formulations of a data-driven approach for solving TSP: supervised training based on the input graph and the ground truth or reinforcement learning based on the input graph and training of the model to minimize the predicted tour cost as done by Dai et al.~\cite{COoverGraphs17}. The authors explore the former and show that the GNN model learns to solve small TSP instances approximately. 
	
	Joshi et al.~\cite{GCNforTSP19} build on top of the approach proposed by Nowak et al.~\cite{nowak2017note}. Specifically, the authors introduce a deep learning model based on GCN for approximately solving TSP instances. The GCN model is fed with a 2D graph. It extracts relevant node and edge features and, as in \cite{nowak2017note}, it directly outputs an adjacency matrix with probabilities for the edges to be part of the TSP tour. The heat-map of edge probabilities is converted into a valid solution using a post-hoc beam search technique \cite{GCNforTSP19}. This approach is different from \cite{PointerNets15}, \cite{NeuralCOwithRL17} and \cite{AttentionVRP19}, where one node is selected at each decoding step (autoregressive approaches). The learning of the GCN model is supervised with pairs of TSP instances and solutions produced by the Concorde TSP solver \cite{applegate2006traveling}.
	The search techniques explored by Joshi et al. \cite{GCNforTSP19} are greedy (the edge with the highest probability is chosen), beam search and beam search with the shortest tour. The beam search is a limited-width breadth-first search: the $b$ edges with the highest probability among the node's neighbors are explored and the top $b$ partial tours are expanded at each stage until all nodes of the graph are visited; the solution is the tour with the highest probability.

	Prates et al. \cite{GNNforDecisionTSP19} solve the decision variant of the TSP (does graph $G$ admit a Hamiltonian path with cost less than a predetermined value?) with deep learning. Specifically, a 
	GNN is used to embed each node and each edge to a multidimensional vector. The model performs as a message-passing algorithm: edges (embedded with their weights) iteratively interchange `messages' with connected nodes. At termination, the model outputs whether or not a route, subject to a desired cost (that is, less than a predefined constant), exists. The training is performed with dual examples: for each optimal tour cost, one decision with smaller and another with larger than the optimal target cost are generated.

	Lemos et al. \cite{lemos2019graph} solve the decision version of the graph coloring problem (does a graph $G$ accept a $C$-coloring?) by training a 
	GNN through an adversarial procedure. The model rests on the idea of message passing as in the decision version of the TSP~\cite{GNNforDecisionTSP19}. Nodes and edges are embedded into high-dimensional vectors, which are updated through the interchange of information with adjacent nodes. The model also produces a global graph embedding for each color. A  node-to-color adjacency matrix  relates each color to all nodes of the graph so that initially any node can be assigned to any color. At termination of the iteration of messages  between adjacent nodes as well as between nodes and colors, the final binary answer is obtained with node voting.
	
	Similar to the latter two approaches, Selsam et al.~\cite{SATsolver19}  design a method based on the idea of message passing. Specifically, Selsam et al. develop a novel MPNN trained as a classifier to predict satisfiability of a SAT problem. The problem is first encoded as an undirected graph (where literals and clauses are represented as nodes, and edges connect literals with clauses they appear in). Then,  the vector space embedding for each node is refined through the iterative message passing procedure \cite{gilmer2017neural}. The proposed neural SAT solver is given a single bit to indicate the satisfiability of  the problem. Furthermore, the model is an end-to-end solver: the solution can be decoded from the network activations. Selsam et al. stress that their model can be applied to arbitrary problems of varying size.

	\subsubsection{Reinforcement learning}
	\label{subsubsec:GNN-DRL}
	
	Dai et al. \cite{COoverGraphs17} solve the TSP, MVC and MaxCut problems with a devised by them novel framework, which is used by later approaches as the main baseline. Specifically, the authors exploit the graph structure of the problem by adopting a deep learning graph embedding network---structure2vec (S2V) \cite{dai2016discriminative}---which captures the relevant information about each node by considering node properties as well as node neighborhood. Dai et al. use a combination of a version of the Q-learning algorithm and fitted Q-iteration \cite{riedmiller2005neural} to learn a greedy policy parametrized by the graph embedding network. At each step, the graph embeddings are updated with new knowledge about the usefulness of each node to the final value of the objective function. The greedy algorithm builds a feasible solution by consecutively incorporating nodes.

	Abe et al.~\cite{abe2019solving} follow the architecture proposed by Dai et al. \cite{COoverGraphs17} in that they rely on reinforcement learning, which is in contrast to Li et al. \cite{COGNNGuidedTreeSearch18}, who apply a supervised learning and tree search.  Different from earlier approaches, Abe et al. replace the Q-learning with an extended AlphaGO Zero \cite{silver2017mastering} motivated approach applicable to combinatorial optimization problems. In essence, AlphaGo algorithms are trained with self-play, which alternates between simulation and play applying a Monte Carlo tree search. At each episode, the improved policies produced by the Monte Carlo tree search and the results generated by the game are used to train the neural network, which in turn improves the estimates of the policy and state-value functions, which are used by the tree search in the next step. Abe et al. extend the AlphaGo to the Markov decision process formalization of the MVC, MC, and MaxCut problems, namely the input graph instances are of different sizes (in contrast to the game settings in \cite{silver2017mastering}) and the solution to the problem is not binary (by contrast to \cite{silver2017mastering}). The former is addressed by introducing GNNs and the latter by normalized rewards. Different GNNs are employed and examined in \cite{abe2019solving}: structure2vec~\cite{dai2016discriminative}, GCN, and graph isomorphism network \cite{xu2018powerful}, which is a type of MPNN.

	\section{Performance}
	\label{sec:discuss}
	
	With the present section, we aim to quickly orient the reader within the realm of the performance of existing machine learning solutions to combinatorial optimization problems. 
	The categorization we make is based on the surveyed combinatorial optimization problems on graphs (summarized in the Appendix).   
	The taxonomy is introduced to facilitate the comparison and analysis of the different methods and to foster conditions for discovering relevant trends. 
	The surveyed publications offer performance comparisons to: 1)~operations research baselines, different heuristics and/or exact solvers and  2)~other machine learning models designed for solving the same problems. 
	Table~\ref{tab:COPcategory} presents the surveyed contributions at a high level of abstraction by synthesizing rather than listing the full set of experimental results reported in the corresponding papers. 
	In the surveyed literature, the performance of the proposed machine learning methods is evaluated in terms of optimality, generalization, and run time (we discuss these and other aspects of the machine learning approaches in Section~\ref{subsec:analysis}). However, in the surveyed literature, there is no uniform measure of the optimality gap. Therefore, in Table~\ref{tab:COPcategory} we provide an overview per contribution, mainly in the form of comparison to other solutions, whereas in the following subsections we incorporate further details. Whenever extensive simulations results are available, we refer the reader to the corresponding reference for details.  Table~\ref{tab:COPcategory} can be used for multiple purposes: to get an overview of the existing machine learning approaches for combinatorial optimization problems on graphs; to detect the deficiencies and advantages of the proposals (we discuss them in the last subsection, \ref{subsec:analysis}); to know in what setting these were evaluated; or to choose the best method. 
	We do not indicate the best performers as the definition of `best' depends on the criterion, which is determined by the objective and the particular problem instance at hand. The criterion could be the run time, optimality of results, trade-off between the two, generalization capability in terms of size of the problem instance, or generalization to different variations of the problem. Therefore, the reader may use Table~\ref{tab:COPcategory} from their individual interest and perspective. 
	Before beginning our discussion with the most extensively researched combinatorial problem, the TSP, we reflect  on one more relevant aspect below.
	
	In Section~\ref{sec:MotivationMLCOG}, we discussed how the focus on computational complexity has evolved through the years, with an increasing emphasis on the theoretical bounds and practical aspects in tandem. Notably, for the surveyed machine learning for combinatorial optimization on graphs contributions, the complexity in terms of Big $\mathcal{O}$ notation has been evaluated only by Dai et al.~\cite{COoverGraphs17} (who clarify that their machine learning architecture has polynomial $\mathcal{O}(k|E|)$, $k \leq |V|$ time-complexity), Vinyals et al.\cite{PointerNets15} (who state that their pointer network implements an $\mathcal{O}(n^2)$ algorithm) and Nowak et al.~\cite{nowak2017note} (who mention that their supervised network has time-complexity no larger than $\mathcal{O}(n^2)$). Most of the remaining 14 papers evaluate  run times. However, as noted by Kool et al.: ``Run times are important but hard to compare: they can vary by two orders of magnitude as a result of implementation (Python vs C++) and hardware (GPU vs CPU).''~\cite{AttentionVRP19}, p.6. Some researchers, such as Prates et al.\cite{GNNforDecisionTSP19}, Lemos et al.\cite{lemos2019graph} or Selsam et al.\cite{SATsolver19},  do not report run times at all. 
	
	In addition to Section~\ref{sec:MotivationMLCOG} and the Appendix (which include complexity discussion and results, respectively), for the reader, we provide references that offer insights  into the computational complexity of some machine learning structures. Specifically, in the context of attention models, Vaswani et al.~\cite{attentionIsAll} compare self-attention to recurrent and convolutional layers.  The comparison is in terms of computational complexity per layer, amount of computation that can be parallelized, and the path length between long-range dependencies in the network (see Table 1 and Section 4 in \cite{attentionIsAll}). Vaswani et al.~\cite{attentionIsAll} show that in terms of computational complexity, attention layers are faster as long as the sequence length is smaller than the representation dimensionality. Liu et al.~\cite{liu2020efficient} discuss the computational complexity of GNNs and GCNs, clarifying that graph convolutional methods require the quadratic graph size for each layer. Spectral GCNs have a worst time complexity $\mathcal{O}(n^3)$, where $n$ denotes the number of graph nodes \cite{liu2020efficient}. For spectral-free methods, the lower bound on time complexity is still $\mathcal{O}(n^2 d)$, where $d$ is the number of channels \cite{liu2020efficient}. Similar time-complexity aspects of GCN are discussed in detail by Zhang et al.~\cite{zhang2019graph}. The time-complexity of recurrent GNNs and GCNs is summarized by Wu et al.~\cite{GNNSurvey} (see Table III therein). The time-complexity of graph autoencoders is discussed by Zhang et al.~\cite{zhang2018deep}. Strehl et al.~\cite{strehl2006pac} clarify that the evaluation of reinforcement algorithms involves in addition to the space-complexity (amount of memory required by an algorithm) and time-complexity (amount of operations executed by the algorithm), the sample-complexity, which measures the amount of experience needed for the algorithm to learn to operate (close to) optimally. The amount of experience is measured in terms of time steps. The sample-complexity of general reinforcement learning is analyzed  by Lattimore~\cite{lattimore2013sample}, and a general treatment is given by Arulkumaran et al.~\cite{8103164}.

	\begin{table*}[!ht]
		\begin{center}
			\begin{tabular}{|l|l|l|c|c|c|l|}
				\hline
				\textbf{COP} &\textbf{Reference} & \textbf{Machine} & \textbf{Learning} & \textbf{Training} & \textbf{Testing} & \textbf{Performance summary:}  \\
				&	& \textbf{learning}& \textbf{structure}&\textbf{graph / dataset}&\textbf{graph / dataset}& \textbf{optimality, generalization \& speed} \\
				\hline
				&Vinyals et al.\cite{PointerNets15}&supervised & pointer network	 & $[5, 20]$ & $\{5, 10, 20\}$ & close to exact solvers / heuristics  \\\cline{2-7}
				&Bello et al.\cite{NeuralCOwithRL17}&reinforcement& pointer network &$\{20, 50\}$	&$\{20, 50\}$	& more optimal than \cite{PointerNets15} 	\\\cline{2-7}
				&Dai et al.\cite{COoverGraphs17}&reinforcement&structure2vec&$\leq[200, 300]$&$\leq$[1k, 1.2k]& more optimal than \cite{NeuralCOwithRL17}, generalizes\\	\cline{2-7}
				&Deudon et al.\cite{deudon2018learning}&reinforcement&Transformer&$\{20, 50\}$	&$\{20, 50, 100\}$	&  more optimal than \cite{NeuralCOwithRL17}, generalizes\\\cline{2-7}
				&Nazari et al.\cite{RLforVRP18}&reinforcement& attention &$\{20, 50, 100\}$&$\{20, 50, 100\}$& as optimal as \cite{NeuralCOwithRL17}, \\ 
				& & & & & & but $\geq55.9\%$ faster than \cite{NeuralCOwithRL17} \\ \cline{2-7}
				TSP	&Kool et al.\cite{AttentionVRP19}&reinforcement&Transformer& $\{20, 50, 100\}$&$\{20, 50, 100\}$& less optimal, faster than Concorde\cite{applegate2006traveling}; \\
				& & & (GAN) & & & smaller optimality gap than  \cite{PointerNets15},\cite{NeuralCOwithRL17},  \\
				& & & & & & \cite{COoverGraphs17}, \cite{nowak2017note}, \cite{deudon2018learning}; \\ 	
				& & & & & & generalization drops quickly\\ \cline{2-7}	
				&Ma et al.\cite{ma2019combinatorial}&(hierarchical)&enhanced&$\{20, 50\}$& $\{20, 50\}$&smaller optimality gap than  \cite{NeuralCOwithRL17}, \cite{COoverGraphs17} \\
				&									&reinforcement &pointer network&	&	& larger optimality gap than \cite{AttentionVRP19}\\\cline{5-7}		
				&	&& &  & & smaller optimality gap than \cite{NeuralCOwithRL17},\cite{AttentionVRP19}; \\
				&	&& & 50 &$\{250, 500, 1k\}$ & generalizes \\\cline{2-7}
				&Wu et al.\cite{wu2019learning}&reinforcement&Transformer&$\{20, 50, 100\}$ & $\{20, 50, 100\}$ &  outperforms \cite{AttentionVRP19} for long test time \\\cline{2-7}		
				&Nowak et al.\cite{nowak2017note}&supervised&GNN& 20 & 20 & slightly less optimal than \cite{PointerNets15} \\\cline{2-7}
				&Joshi et al.\cite{GCNforTSP19}&supervised& GCN & $\{20, 50, 100\}$ & $\{20, 50, 100\}$ &  compared to \cite{PointerNets15}, \cite{NeuralCOwithRL17}, \cite{COoverGraphs17}, \cite{AttentionVRP19}  \\ \cline{2-7}
				&Prates et al.\cite{GNNforDecisionTSP19}& supervised & GNN	& $\{20, 40\}$ & $[20, 80]$ & more optimal than NN and SA; \\
				& & & & & &generalization drops quickly\\ 
				\hline
				
				VRP 	&Nazari et al.\cite{RLforVRP18}&reinforcement& attention &$\{50, 100\}$&$\{50, 100\}$& $61\%$ shorter tours than OR-tools\cite{teamor}\\\cline{2-7}
				\&var.&Kool et al.\cite{AttentionVRP19}&reinforcement&Transformer& $\{20, 50, 100\}$&$\{20, 50, 100\}$& smaller optimality gap than \cite{RLforVRP18} \\
				\hline
				
				&Dai et al.\cite{COoverGraphs17}&reinforcement&structure2vec&$\leq[200, 300]$&$\leq$[1k, 1.2k]&small optimality gap, generalizes \\	\cline{2-7}
				MaxCut	&Barrett et al.\cite{barrett2019exploratory}&reinforcement&MPNN&$[20, 200]$ & 500, 2k& outperforms \cite{COoverGraphs17}, generalizes \\\cline{2-7}
				&Abe et al.\cite{abe2019solving}&reinforcement&GNN&$[40, 50]$ &synthetic \& real& compared to \cite{COoverGraphs17}; generalizes\\
				\hline		
				
				&Dai et al.\cite{COoverGraphs17}&reinforcement&structure2vec&$\leq[400, 500]$&$\leq$[1k, 1.2k]&small optimality gap, generalizes\\		\cline{2-7}
				&Li et al.\cite{COGNNGuidedTreeSearch18}&supervised& GCN & synthetic\cite{grosso2008simple}  & social\cite{leskovec2014snap} \& & outperforms \cite{COoverGraphs17}\\
				&  &  &  &  & citation\cite{sen2008collective}  & \\\cline{2-7}
				MVC	&Mittal et al.\cite{LearningHeuristicsLargeGraphs19}&supervised& GCN & 1k & $\leq 2$k & more optimal than \cite{COoverGraphs17}, less optimal\\ 
				& & & & & &  than \cite{COGNNGuidedTreeSearch18} but faster than \cite{COGNNGuidedTreeSearch18} \\\cline{2-7}
				&Abe et al.\cite{abe2019solving}&reinforcement&GNN&$[80, 100]$ &synthetic \& real& compared to \cite{COoverGraphs17}; generalizes\\
				\hline
				
				MC	&Li et al.\cite{COGNNGuidedTreeSearch18}&supervised& GCN & synthetic\cite{grosso2008simple} & synthetic\cite{xu2007random}& $62.5\%$ solved, outperforms \cite{COoverGraphs17}\\	\cline{2-7}
				&Abe et al.\cite{abe2019solving}&reinforcement&GNN&$[80, 100]$ &synthetic \& real& compared to \cite{COoverGraphs17}; generalizes\\
				\hline
				
				&Li et al.\cite{COGNNGuidedTreeSearch18}&supervised& GCN &  & SAT Competition & $100\%$ solved, slower than \\
				& &	& & & \cite{balyo2017sat}			&Z3 solver\cite{de2008z3} \\\cline{6-7}
				MIS& & &   &  & social\cite{leskovec2014snap} \&& best results ($\#$ solutions found)  \\
				& & &   &  & citation\cite{sen2008collective}& among all baselines including \cite{COoverGraphs17}\\
				\hline
				
				MCP	&Mittal et al.\cite{LearningHeuristicsLargeGraphs19}&supervised& GCN & 1k & $\leq 2$k & more optimal than \cite{COoverGraphs17}, \cite{COGNNGuidedTreeSearch18}\\
				\hline
				
				GC	&Lemos et al.\cite{lemos2019graph}& supervised	&  GNN  & $[40, 60]$ & $\leq 561$ & varying accuracy, generalizes\\
				\hline
				
				SAT	&Selsam et al.\cite{SATsolver19} &  supervised	&  MPNN & $40$ & $[40, 200]$ & generalizes to larger instances \\
				\hline
				
				KP 	&Bello et al.\cite{NeuralCOwithRL17}&reinforcement&pointer network&$\{50, 100, 200\}$&$\{50, 100, 200\}$ & optimal \\
				\hline
			\end{tabular}
		\end{center}
		\caption{Combinatorial optimization problems (COPs): machine learning contributions that solve them, with their performance summarized. Abbreviations according to Appendix and Section~\ref{sec:theory}. 
			NN stands for nearest neighbour, SA for simulated annealing.} 
		\label{tab:COPcategory}
	\end{table*}

	\subsection{Travelling Salesman Problem}
	\label{subsec:TSP}
	
	Vinyals et al.~\cite{PointerNets15} train their pointer network with supervised learning. For generating training samples, the coordinates of the nodes are sampled from a Cartesian unit square, which is a commonly followed procedure in the  studies discussed here. The model is supervised with exact solutions for graph instances with a total number of nodes $n$ at most 20 and with approximate solutions for larger graphs. Testing for the same number of nodes as that during training $n=\{5, 10, 20\}$ produces optimal or very close to optimal results. The pointer network generalizes for $n=25$, but the solution quality deteriorates quickly and breaks at $n=40$ \cite{PointerNets15}. 
	
	In short, one of the major contributions of Vinyals et al. is that they develop a model with which they demonstrate that ``a purely data-driven approach can learn to approximate solutions to intractable problems'' ~\cite{PointerNets15}, p. 2. The main downside of the proposed model from TSP performance perspective is its limited applicability. It does not generalize to larger instances of interest and it needs to be supervised. Indeed, the difficulty of obtaining exact solutions, as remarked by the authors, restricts the practical use of the model since the quality of the labels is a prerequisite for obtaining optimal solutions at test time.

	Bello et al.~\cite{NeuralCOwithRL17} show that a pointer network trained with reinforcement learning surpasses the performance of the supervised pointer network of Vinyals et al.~\cite{PointerNets15}. The training and testing is conducted on graph instances with  $n=\{20, 50, 100\}$ nodes. Comparisons are also made to baselines: the results reported by the authors are inferior to the Concorde~\cite{applegate2006traveling} solver, which produces provably correct solutions, but the greedy RL pretraining versions of their model are much faster than the solver. Furthermore, optimality of the results can be traded off for run time with the active search version of their model (Bello et al. report more extensive experimental results with the different modalities of the approach  in \cite{NeuralCOwithRL17}).
	
	Deudon et al. \cite{deudon2018learning} train the enhanced framework of Bello et al.~\cite{NeuralCOwithRL17} on graph instances with  $n=\{20, 50\}$ nodes and test on the same size as well as on graphs with $n=100$. The hybrid heuristic of Deudon et al. outperforms the framework of Bello et al. for all of these cases.  The authors also report performance results compared to Google OR-Tools VRP engine~\cite{teamor} and Concorde~\cite{applegate2006traveling} on the same test set: the run  time of their framework is worse than the former, close to the latter, and significantly better (shorter run time) than that of Bello et al.
	
	The S2V-DQN framework developed by Dai et al.~\cite{COoverGraphs17} can learn to solve TSP of a larger size. The authors evaluate the quality of the obtained solutions by an approximation ratio, defined by $\mathcal{R}(S_I,P_I) = \max(\frac{opt(P_I)}{c(h(S_I))}, \frac{c(h(S_I))}{opt(P_I)}$, where $c(h(S_I))$ is the objective value of the solution $S_I$ and $opt(P_I)$ is the best known solution of problem instance $P_I$. The S2V-DQN architecture is compared to several baselines as well as to the model proposed by Bello et al.~\cite{NeuralCOwithRL17}. When testing and training is conducted over the same distribution and for graphs with nodes up to $n=300$, S2V-DQN is very close to optimal ($\mathcal{R}(S_I,P_I) \in [1, 1.1)$), and outperforms \cite{NeuralCOwithRL17} as well as most of the considered baselines.
	Dai et al. demonstrate that the  graph embedding architecture allows the same set of model parameters to be used and hence the model can be trained and tested on different graphs, which essentially leads to a model that can  generalize to larger graphs. Specifically, when trained on graphs with $n=[50, 100]$ nodes, the S2V-DQN algorithm attains $\mathcal{R}(S_I,P_I)=1.0730$  for $n=[50, 100]$ and  $\mathcal{R}(S_I,P_I)=1.1065$ for $n=[1000, 1200]$. Li et al. report in \cite{COGNNGuidedTreeSearch18} S2V-DQN results for even larger graphs, namely that  S2V-DQN performance  deteriorates for graphs with more than 10,000 (10k) nodes.
	
	Nazari et al.~\cite{RLforVRP18} train and test for TSP instances of size $n=\{20, 50, 100\}$ drawn from the same distribution and compare the results from their greedy algorithm (the node with the highest probability during the decoding step is chosen as the next city to be visited) with those obtained by Bello et al.~\cite{NeuralCOwithRL17}. Their results are very close to but suboptimal to those reported by Bello et al. \cite{NeuralCOwithRL17}. Nonetheless, the approach of Nazari et al. (which is an extension of that of Bello et al.) is much faster with time savings in the order of $55.9\%$ or more for the examined TSP instances. 
	
	The model of Kool et al.~\cite{AttentionVRP19} is trained and tested on graph instances with $n=\{20, 50, 100\}$ nodes and the optimality of its solutions is compared to \cite{PointerNets15}, \cite{NeuralCOwithRL17}, \cite{COoverGraphs17}, \cite{nowak2017note}, and \cite{deudon2018learning}. Kool et al. obtain  smaller optimality gap (when compared to Concorde) than the aforementioned studies as well as existing heuristics approaches. Generalization capacity is tested on instances with up to $n=125$ nodes, but as remarked by the authors the performance gap (measured with regard to Gurobi~\cite{optimization2014inc}) increases quickly when the gap between train and test sizes is increased (see Appendix B in \cite{AttentionVRP19}).
	
	Ma et al.~\cite{ma2019combinatorial} train and test their graph pointer network on small TSP instances with $n=\{20, 50\}$ nodes for which the reported results are less optimal than those of Kool et al. \cite{AttentionVRP19} yet better than those of Bello et al. \cite{NeuralCOwithRL17} and Dai et al. \cite{COoverGraphs17}. Note that the results reported by Ma et al. for the models of Bello et al.~\cite{NeuralCOwithRL17} and Dai et al.~\cite{COoverGraphs17} are not consistent with those reported in \cite{COoverGraphs17}, where Dai et al. demonstrate that their model obtains smaller optimality gap than Bello et al. for all problem instances. The generalization capacity of the hierarchical reinforcement learning framework is tested for instances with up to $n=1000$ nodes, when the training is conducted on graphs with $n=50$. When compared to Concorde \cite{applegate2006traveling}, Ma et al. report a smaller approximate gap than that obtained with the approaches of Bello et al. \cite{NeuralCOwithRL17} and Kool et al. \cite{AttentionVRP19}, but larger than that of Dai et al. \cite{COoverGraphs17}. With the local 2-opt~\cite{croes1958method} search the reported results are improved. With that heuristic added, Ma et al. \cite{ma2019combinatorial} report results with smaller approximate gap to the optimal solution (computed by Concorde \cite{applegate2006traveling}) when compared to Dai et al. \cite{COoverGraphs17}. Furthermore, the test run time is considerably smaller than that of Concorde. 
	
	Wu et al.~\cite{wu2019learning}, who learn improvement heuristics policy from random initial solutions (of usually poor quality), demonstrate that their method generalizes to unseen initial solutions and that the quality of the solution obtained by their improvement heuristic (based on 2-opt~\cite{croes1958method}) policy does not depend on the initial solution quality. When compared to the solutions of the exact solver Concorde~\cite{applegate2006traveling}, the optimality gap is practically 0 for graph instances with $n=20$ nodes, less than $1\%$ for $n=50$ and less than $3.5\%$ for $n=100$ (the exact values depend on the training time limit $T$). For $T=3,000$ time steps during testing ($T=200$ during training), the method produces results with a smaller optimality gap (measured with respect to the results obtained with Concorde) than the method of Kool et al.~\cite{AttentionVRP19} with sampling. Wu et al.~\cite{wu2019learning} also show that for harder problems (i.e., larger problem instances), their method is more efficient than sampling \cite{AttentionVRP19} in terms of execution time. The learned improvement heuristics are compared to conventional ones and are shown to be better in terms of optimality gap and potentially in terms of computation time (the authors indicate that a direct comparison with regard to the latter criterion is not straightforward, see \cite{wu2019learning} for further details).
	
	Nowak et al.~\cite{nowak2017note} train and test on small TSP instances with $n=20$ nodes and obtain results that are slightly less optimal than that produced in \cite{NeuralCOwithRL17} and by another common baseline heuristic. The authors recognize that the major disadvantage of their data-driven approach is the ``need for expensive ground truth examples, and more importantly, the fact that the model is imitating a heuristic rather than directly optimizing the TSP cost'' \cite{nowak2017note}.

	The GCN approach of Joshi et al.~\cite{GCNforTSP19}, when using a greedy search, produces a tour length with smaller optimality gap than \cite{PointerNets15}, \cite{NeuralCOwithRL17}, and \cite{COoverGraphs17} for instances of size $n=\{20, 50\}$ but a larger one for $n=100$. 
	The optimality gap is defined as the average percentage ratio of the predicted tour length $l^{TSP}$ relative to the optimal solution $\hat{l}^{TSP}$ over 10k instances: $(1/m) \sum_{i=1}^{m}(l^{TSP}_m / \hat{l}^{TSP} - 1)$. When compared to the attention model with reinforcement learning of Kool et al.~\cite{AttentionVRP19}, the GCN model proposed by Joshi et al.~\cite{GCNforTSP19} has a larger optimality gap measured by the TSP tour length and is slower for all studied graph sizes. When combined with beam search, GCN is faster than the model of Kool et al.~\cite{AttentionVRP19}. In contrast, the GCN model with beam search and shortest path heuristic produces minimal optimality gap at the cost of much longer evaluation times. One of the advantages of the GCN approach is that it can be parallelized for GPU computation \cite{GCNforTSP19}, which decreases the inference time and improves scalability to larger TSP instances. In contrast,  autoregressive models (such as \cite{COoverGraphs17}) are less amenable to parallelization since the decoding process is sequential \cite{GCNforTSP19}. The authors also note that reinforcement learning does not require labels but assert that it is less sample efficient and hence more computationally expensive than supervision. Furthermore, upon a sufficient amount of training data, supervision models can outperform reinforcement learning techniques. However, for larger graphs (beyond hundreds of nodes) this advantage becomes less prominent. Therefore, Joshi et al. reason that extending their approach with reinforcement learning is a natural line of research (with transfer learning from smaller instances) for making it scalable to real-world problems.

	\subsection{Vehicular routing problem}
	Nazari et al.~\cite{RLforVRP18} show that their reinforcement solution is competitive with state-of-the-art VRP heuristics. Compared to the Google OR-Tools VRP engine~\cite{teamor} (among the best open-source VRP solvers according to \cite{RLforVRP18}), their method outputs shorter tours in roughly $61\%$ of the cases for the capacitated VRP instances with $n=\{50, 100\}$ customers. The authors suggest that their model is robust to problem changes (in demand value or location of a customer) because the model can automatically adapt the solution to these. Another appealing property (split delivery) of the model is that when more than one vehicle can deliver supplies to a single node, the  framework can outperform the single vehicular deliveries with no additional cost or hand-engineered rules. Another set of experimental results shows the log of the ratio of solution run times to the number of customer nodes. It remains almost constant for their approach, whereas for classical heuristic methods it increases faster than linearity with the number of nodes. 
	Based on this observation, Nazari et al. conclude that their method scales. For the stochastic VRP, the model  produces better results than simple baseline heuristics.

	Kool et al.~\cite{AttentionVRP19} solve TSP and VRP problems as well as several of the VRP variants using attention mechanisms and reinforcement learning. A distinctive feature of the model developed by Kool et al. is that in the experiments it is used to solve four different routing problems (each one with its objective and constraints), which demonstrates the adaptability of the model to different problem settings. Indeed, recall that versatility is one of the envisioned strong aspects of machine learning for combinatorial optimization, namely the ability of a learned model to adapt to different problems instead of devising new heuristics for each possible problem modality. In short, the main advantage of the designed method is that all of the considered problems can be solved with it; moreover, with a single set of hyper-parameters. Otherwise, the problem instances (both for training and testing) are up to $n=100$ nodes. For the capacitated VRP and stochastic VRP, the attention model with reinforcement learning of Kool et al. \cite{AttentionVRP19} produces smaller objective value and importantly smaller optimality gap (to the best value across all models) when compared to the model of Nazari et al.~\cite{RLforVRP18}. In addition, for the prize collecting TSP, Kool et al. obtain results of quality similar to this achieved by Google OR-tools~\cite{teamor}, but Kool et al. model is considerably  faster.

	\subsection{Maximum cut}
	The solution quality of the S2V-DQN framework of Dai et al.~\cite{COoverGraphs17} is evaluated by the approximation ratio defined earlier in Section~\ref{subsec:TSP}. For training and testing, Barab{\'a}si-Albert~\cite{albert2002statistical} graphs are generated. The number of nodes is sampled uniformly at random from the $[200, 300]$ range. The optimal solutions are produced by CPLEX~\cite{cplex2009v12}. The results produced by S2V-DQN are very close to optimality ($\mathcal{R}(S_I,P_I)<1.05$). The framework generalizes to larger instances and for $n \in [1000, 1200]$, the approximation ratio averaged over 1,000 test instances is $\mathcal{R}(S_I,P_I)=1.0038$.
	
	Barrett et al.~\cite{barrett2019exploratory} take as a baseline the S2V-DQN approach of Dai et al.~\cite{COoverGraphs17} for performance evaluation. Barab{\'a}si-Albert~\cite{albert2002statistical} and Erd{\H{o}}s-R{\'e}nyi~\cite{erdHos1960evolution} graphs are generated for training and testing. An approximation ratio $C(s_*)/C(s_{opt})$, where the numerator is the cut value of the approach and the denominator is the optimum, (\textit{optimum} is the best solution obtained by any of the approaches: CPLEX or advanced heuristics, see \cite{barrett2019exploratory} for further details) is used for evaluation. When trained and tested on problem instances of the same size $n \in \{20, 40, 60, 100, 200\}$,  their model  improves on the performance of the S2V-DQN framework. The generalization performance is tested on generated graph instances of up to $n=500$, when the training is performed on $n=40$. When the model is tested on Barab{\'a}si-Albert but trained on Erd{\H{o}}s-R{\'e}nyi (and vice versa), there is some decrease in performance, but in all studied cases the model outperforms S2V-DQN, while the authors state that the computational cost per agent action is similar for the two approaches. On known benchmarks of up to 2000 nodes, the observed performance is (very close to) optimal despite that the model is trained and tested on graphs with different structures. Detailed analysis and a much richer set of performance results are included in \cite{barrett2019exploratory}. We note that the ablations in \cite{barrett2019exploratory} show that the improved performance is due to: 1) the freedom of the agent to reverse its earlier decisions and 2) the provision of the agent with suitable observations and rewards. Moreover, the model has the flexibility to be deployed with different heuristics, which  can bring further improvements. Another result is that taking the best solution across multiple randomly initialized episodes leads to significant performance improvements. 
	
	For training their AlphaGo Zero inspired model, Abe et al. \cite{abe2019solving} generate Erd{\H{o}}s-R{\'e}nyi~\cite{erdHos1960evolution} graphs with $n=[40, 50]$. Testing is conducted on synthetic (Barab{\'a}si-Albert~\cite{albert2002statistical} and Erd{\H{o}}s-R{\'e}nyi) as well as real-world data sets (the latter are from the Network Repository \cite{rossi2015network}). For most of the cases, Abe et al. \cite{abe2019solving} report better performance when compared to S2V-DQN \cite{COoverGraphs17}. This is achieved at the cost of significant increase in computation power and longer training time due to the self-play characteristics of the approach (the use of Monte Carlo tree search). Otherwise, Abe et al.~\cite{abe2019solving} state that their approach is more sample-efficient than S2V-DQN.

	\subsection{Minimum vertex cover}
	Dai et al.~\cite{COoverGraphs17} also solve the MVC problem with their S2V-DQN model in a similar setting as MaxCut. The main difference is that the ranges of the training and test graph sizes are up to $[400, 500]$. The best ratio---practically 1 (results very close to optimality)---for small graphs (with $n \leq 500$) is produced for this problem. The experimental results also show generalization capabilities for graphs with nodes in the range of $[1000, 1200]$,  where the average approximation ratio is $\mathcal{R}(S_I,P_I)=1.0062$.

	Li et al.~\cite{COGNNGuidedTreeSearch18} solve four canonical $\mathcal{NP}$-hard problems: SAT, Maximal
	Independent Set (MIS), MVC, and MC. The experimental results indicate that their approach obtains optimal solutions similar to those of optimized contemporary solvers based on traditional heuristic methods. In addition, the authors conclude that their approach generalizes across datasets as well as graph instances orders of magnitude larger than those used during training. For the MVC problem, the model is tested on various data sets ranging from synthetic problem instances, SATLIB \cite{grosso2008simple} and BUAA-MC \cite{xu2007random} data sets (the latter  includes 40 hard synthetic MC instances), to real data from social, SNAP ~\cite{leskovec2014snap}, and citation \cite{sen2008collective} networks. Best results are reported for all test cases, namely outperforming all baselines including S2V-DQN \cite{COoverGraphs17}. 
	
	Mittal et al.~\cite{LearningHeuristicsLargeGraphs19} train their model on graphs of size 1k. On large data sets, namely Barab{\'a}si-Albert~\cite{albert2002statistical} graphs with at most 500k nodes, the model outperforms S2V-DQN and has similar performance to that of  Li et al. \cite{COGNNGuidedTreeSearch18}, see \cite{LearningHeuristicsLargeGraphs19} for details. Mittal et al. 
	argue that their approach is scalable to billion size networks and is orders of magnitude faster than S2Q-DQN \cite{COoverGraphs17} and the GCN with tree search \cite{COGNNGuidedTreeSearch18} methods. 
	
	Abe et al.~\cite{abe2019solving} consider the S2V-DQN model as a baseline and some chosen heuristics too. Erd{\H{o}}s-R{\'e}nyi~\cite{erdHos1960evolution} graphs with $n=[80, 100]$ nodes are used for training. Depending on the particular implementation of a GNN, the proposed  framework with Monte Carlo tree search has a varied success when compared to S2V-DQN (see Table 1 and Table 2 in \cite{abe2019solving}). Overall, the proposed model generalizes to much larger graph instances according to the results in \cite{abe2019solving}.

	\subsection{Maximal independent set and Maximal clique}
	
	The model of Li et al.~\cite{COGNNGuidedTreeSearch18} is trained on the SATLIB benchmark~\cite{grosso2008simple} and tested on MIS problems from the SATLIB~\cite{grosso2008simple} and SAT Competition~\cite{balyo2017sat} data sets for which $100\%$ of the solutions are found. On the SNAP social networks data sets~\cite{leskovec2014snap} and various  citation networks, the authors report that their model produces the best results among the models tested (including S2V-DQN \cite{COoverGraphs17}). Similarly, the best results for MC problems from the BUAA-MC \cite{xu2007random} data set are obtained. 
	
	Abe et al. \cite{abe2019solving} show results for MIS and MC. The training is conducted on Erd{\H{o}}s-R{\'e}nyi~\cite{erdHos1960evolution} graphs with $n=[80, 100]$ nodes and testing on synthetic and real-world data (as for the other problems considered, namely MVC and MaxCut). For MC they report better results (obtained with different GNN implementations) than those obtained with S2V-DQN. For the MIS, the results are varied (see Table 2 in Appendix D \cite{abe2019solving}). For one instance of MIS, with graph isomorphism network implementation, the authors report a result better than CPLEX. The computational complexity and run time of the framework \cite{abe2019solving} is much higher than that of the baseline S2V-DQN \cite{abe2019solving}, as noted earlier.

	\subsection{Maximum Coverage Problem}  
	
	Mittal et al. \cite{LearningHeuristicsLargeGraphs19} solve the MCP (see the Appendix for definition) on Barabasi-Albert~\cite{albert2002statistical} graphs of at least $n=2$k nodes and compare the performance results to those from  S2V-DQN \cite{COoverGraphs17}) and GCN with three search \cite{COGNNGuidedTreeSearch18}, which their model outperforms. Moreover, they report that their approach is four to five times faster than that of Li et al.~\cite{COGNNGuidedTreeSearch18}.
	The authors attribute this speed-up to the single call to GNN and prediction using Q-learning instead of tree search.

	\subsection{Satisfiability}
	To make the proposed neuro SAT network learn to solve SAT problems, Selsam et al. \cite{SATsolver19} create a distribution SR($n$) over pairs of random SAT problems on $n$ variables. Each pair consist of one satisfiable and another  not satisfiable element. The difference between the two elements is obtained by negating only a single literal occurrence in a single clause. Then, training samples are generated from this distribution. Although the  model does not show competitive results when compared to specialized SAT solvers, the authors demonstrate that such problems can be solved with MPNN. An important observation made by Selsam et al.~\cite{SATsolver19} is that their model solves substantially larger and harder problems than those seen during training.

	\subsection{Graph coloring}
	The GNN model proposed by Lemos et al. \cite{lemos2019graph} can be trained to solve graph coloring problems with up to $82\%$ accuracy and the model can scale to unseen larger instances: the training is conducted on $n \in [40, 60]$ and $c \in [3, 7]$ (where $n$ and $c$ are the number of nodes and colors, respectively), whereas the largest problem test instance is with $n = 561$.

	\subsection{Discussion}
	\label{subsec:analysis}
	
	Two contemporary structures emerge on the surface as the presently most prominent ones for tackling combinatorial challenges of the surveyed kind: attention mechanisms for considering relevant context information, and GNNs for accounting for the inherent graph structure of the problems. Both methods aim at one and the same relevant objective: capturing the (complex) relations between the nodes in the graph. 
	The results summarized in Table~\ref{tab:COPcategory} suggest that both---architectures that employ attention and frameworks that are based on GNNs---obtain close to optimality results. Looking at the generalization performance, results for large graph instances are reported for GNN models (and  they scale): \cite{COoverGraphs17}, \cite{barrett2019exploratory}, \cite{COGNNGuidedTreeSearch18} (reinforcement learning), and \cite{LearningHeuristicsLargeGraphs19} (supervised)), whereas results for equally large problems are not reported for models that apply attention mechanisms.
	On the other hand, supervised learning and deep reinforcement learning seem to be competitive, considering performance results, each coming with their own advantages and shortcomings. Recall that the most outstanding feature for supervised learning is the need for labeled data, which might be challenging considering the complexity of the problems and the quality of the available solutions. However, when training data sets are available, supervised learning often requires less training time than reinforcement learning. The main shortcomings of reinforcement learning approaches are the potential instability and divergence problems,  although there are advancements in overcoming them. We note that the general advantages and limitations of these two learning approaches are known and hence we do not discuss them further (see Sutton and Barto~\cite{SuttonBarto2018} and also Joshi et al.~\cite{GCNforTSP19}). 
	
	It is worth emphasizing that critical enabling factors for applying machine learning and in particular deep learning to combinatorial optimization, are the structure of the problems and the (often vast) availability of data, which is due to the same problem emerging again and again in various practical fields. One primary ambition, when exploring machine learning as a tool for solving combinatorial optimization problems, is to circumvent the need of deep expertise into a particular domain firstly because it might not be available and secondly because it might be difficult and certainly time consuming to acquire such in a reasonable time span. Therefore, the  goal in this context is to automate the learning of useful heuristics. Another important motivation that drives renewed interest and research effort in this field is the need to develop algorithms faster than existing solvers, simply because the new large-scale challenges typical for the present-day require novel and innovative solutions. An even further requirement and therefore a higher goal (considering the existence of optimized solvers) is to make the novel tools produce better quality solutions than available heuristics.
	
	For the machine learning models for combinatorial optimization problems on graphs to be practically applicable and able to overcome some of the limitations of the presently available tools, we identify a handful of relevant challenges that need to be addressed first:
	\begin{itemize}
		\item \textbf{Scalability}. The machine learning models need to perform at the same observed quality level when applied to problems from the same data generating function but of a (usually much) larger size. We have seen examples \cite{COoverGraphs17}, \cite{LearningHeuristicsLargeGraphs19}, \cite{SATsolver19} in which the proposed models can indeed scale to much larger instances according to reported results. However, for the majority of the surveyed publications, the quality of the solution drops quickly when the graph instance is (slightly) increased.
		
		\item \textbf{Adaptability} to perturbation in the problem setting, or adaptability to other problems from the same family. Such a feature is desirable as it will overcome the need to redesign and reoptimize algorithms when the problem changes. We have seen solutions when a machine learning framework was able to adapt to changes in the problem \cite{RLforVRP18} as well as a model that is able to solve different (although from the same class) combinatorial optimization tasks \cite{AttentionVRP19}.
		
		\item \textbf{Generalization}. The ability of the machine learning algorithm to perform well on unseen instances (which inherently involves the first two notions) is of much relevance. Usually, the models are trained under certain setting (environment in the case of reinforcement and labels in the case of supervised), whereas they should perform well on unseen instances (from the same generally unknown distribution)  in order for the models to be of practical interest. Some of the surveyed models show such generalization capacity \cite{COoverGraphs17}, \cite{LearningHeuristicsLargeGraphs19}, and \cite{SATsolver19}.
		
		\item \textbf{Execution time}. The run time is one of the most relevant factors together with the  model performance for the practical implementation of the algorithms, as many applications require quick decisions or within reasonable time frames. We have seen examples,\cite{NeuralCOwithRL17} and \cite{AttentionVRP19}, where the performance of the developed algorithms is close to that of optimal solvers but (significantly) faster.
		
		\item \textbf{Automation}. Another desirable property of machine learning models is to automate the process of meeting the above requirements, namely automatically scale to larger problem instances without the need to re-devise, or manually modify the parameters, automatically adapt to new modifications of the problem definition, automatically generalize to unseen instances without the need to re-train (completely), automatically improve performance with experience.
	\end{itemize}

	Machine learning for solving combinatorial optimization problems on graphs has seen a surge in the development of novel models for solving these challenging tasks. The main motivating factor behind this trend is the broad applicability that such methods would have in diverse fields, where combinatorial problems need to be routinely solved. The first machine learning publications from the past five years were mainly concerned with demonstrating that machine learning methods can solve such hard problems. Later approaches have set more ambitious goals, which include developing models that can generalize beyond the training examples.  Most recent contributions have proposed methods that produce similar quality of results compared to well-engineered solvers but are faster. The combinatorial optimization area attracts more and more interest, evidenced by the increasing number of research contributions on the subject. Overall, in the context of the long history of the operations research field, the progress made by the machine learning community in the past few years is substantial. The currently existing models might not be applicable to an as extensive range of practical domains and specific tasks as they are foreseen to be for the reasons mentioned above (scalability, adaptability, generalization, execution time). However, considering the pace with which advances are made in this field and the results achieved with the models already developed in these few years, we could expect practically applicable machine learning models to emerge from this active research field in the foreseen future. 
	
	The synergy  between machine learning and traditional methods is discussed by Lombardi and Milano~\cite{lombardi2018boosting}, and by Bengio et al.~\cite{bengio2018machine}. We remark that it is expected that the combined power of these two fields can boost performance and speed up the process of designing optimal yet practically applicable  solutions.

	\section{Networking Applications}
	\label{sec:COinNet}
	
	\label{sec:netw-intro}
	Large-scale communication networks require efficient methods for automated network operations and management of infrastructure resources. Network operators and service providers face challenges in dealing with conflicting resource management optimization objectives while aiming to meet the expectations of energy-efficient delivery of low-latency services to billions of users. Greedy algorithms or heuristics can produce good solutions within reasonable time frames for problems with few constraints, but often require a tailored approach to the problem instance based on deep human knowledge about the domain. 
	
	Given our focus on learning network structures, we seek to relativize our findings to practice. We orient this part of the survey to a generic assessment of machine learning applied for solving combinatorial optimization problems in the networking domain in the context of resource management. In general, applied machine learning for solving complex resource management problems has gained increasing interest in the networking community. Rather than delving further into theoretical analyses of worst-case computational complexity of different solutions to well-known combinatorial optimization problems, we assess generalizability, adaptability, and scalability aspects (Section~\ref{sec:discuss}) of interesting candidate solutions developed over the past five years. Specifically, the solutions that could pertain to open questions in live massive network maintenance have directed our criteria for inclusion of research efforts in this section. The discussed research contributions solve representative instances of resource management problems by applying one or a combination of methods presented in Section~\ref{sec:MLforCOP}, but do not constitute an exhaustive analysis of applied machine learning for the networking domain.
	
	\begin{table*}
		\centering	
		\begin{tabular}{|l|l|l|l|c|l|}
			\hline
			\textbf{Problem} & \textbf{Sub-problem} & \textbf{Reference} & \textbf{Year} & \textbf{Learning} & \textbf{Approach} \\
			\hline
			Routing& Network modeling & Rusek et al.~\cite{rusek2019unveiling} & 2019 & supervised & GNN based on MPNN
			\\\cline{2-6}
			\ref{sec:netw-routing}& Routing policy optimization & Almasan et al.~\cite{almasan2019deep} & 2020 & reinforcement & GNN + Q-learning based
			\\ \hline
			Scheduling & Cluster scheduling & Mao et al.~\cite{Mao2018} & 2018 & reinforcement & GNN + custom policy network
			\\ \cline{2-6}
			\ref{sec:netw-scheduling} & Wireless edge content sched. & Wang et al.~\cite{Wang2020a, Wang2019} & 2020, 2019 & reinforcement & A3C 
			\\ \cline{2-6}
			& Multi-path TCP packet sched.  & Zhang et al.~\cite{Zhang2019} & 2019 & reinforcement & LSTM + deep Q-network 
			\\ \hline
			Resource&VNF deploymet & Mijumbi et al.~\cite{Mijumbi2017, Mijumbi2016} & 2017, 2016 & supervised & GNN based on MPNN 
			\\ \cline{2-6}
			allocation& Edge network slicing & Liu et al.~\cite{Liu2020} & 2020 & reinforcement & Deep Q-network based 
			\\ \cline{2-6}
			\ref{sec:netw-resourceallocation}& Fog computing & Mseddi et al.~\cite{Mseddi2019} & 2019 & reinforcement & Deep Q-network based
			\\ \hline
			
		\end{tabular}
		\caption{Overview of illustrative approaches applying machine learning for solving CO problems. Detailed description is found in subsections \ref{sec:netw-routing} Routing, \ref{sec:netw-scheduling} Scheduling and \ref{sec:netw-resourceallocation} Resource allocation.}
		\label{tab:NetworkingOverview}
	\end{table*}
	
	\subsection{Routing}
	\label{sec:netw-routing}
	
	The routing problem, which consists of finding and selecting a (set of) path(s) between a source and a destination subject to some constraints (such as delay and required capacity), arises in many different domains. It is a classical resource allocation problem in communication networks too. We illustrate here how it has been solved in the networking domain with contemporary machine learning concepts. We begin with a discussion on network modeling. Although it is not a combinatorial problem, network modeling is a relevant subtask of the routing optimization task. 
	Consider, for instance, a routing objective as follows:  route the traffic demands subject to delay constraints. A model for estimating a delay metric is needed then.

	 \begin{table*}[!ht]
		\begin{center}
			\begin{tabular}{|l|l|l|l|l|l|}
				\hline
				\multicolumn{6}{|c|}{\textbf{Routing}}\\
				\hline
				\textbf{Problem} &\textbf{Reference} & \textbf{Learning} & \textbf{Approach} & \textbf{Performance summary} & \textbf{Remarks}   \\
				\hline
				Network&Rusek et al.\cite{rusek2019unveiling}&supervised& MPNN &Model learns relation between   &Flow paths must be known a priori.\\
				modeling&&&& topology, routing, and input traffic. &  Assumes same capacity for all links.\\
				&&&&& Generalization: trained on 14-node \\
				&&&&& NSFNet \cite{nettopo}, tested on Geant2 with 24 \\
				&&&&& and GBN with 17 nodes \cite{knight2011internet}: \\
				&&&&& generalizes to these two. \\ 	
				\hline
				Routing&Almasan et al.\cite{almasan2019deep}&reinforcement& MPNN &Outperforms baseline learning &Trained on  a single 14-node NSFNet\\
				policy&&&&method in $80\%$ cases. &  topology \cite{nettopo}. Tested on 136 topologies\\
				optimization&&&& Robust to link failures. & \cite{knight2011internet} with nodes $n=[5,50]$: generalizes\\
				&&&&& to these unseen topologies. \\ 			
				\hline
			\end{tabular}
		\end{center}
		\caption{Routing problem: machine learning approach and performance} 
		\label{tab:CoinRout}
	\end{table*}
	
	\subsubsection{Network modeling}

	The analytical modeling of communication networks is central to network management and optimization. To render the resulting models mathematically tractable many simplifying assumptions concerning arrival processes and queueing times, for instance, are often made. Although useful in earlier generation of networked systems, these models do not seem practically applicable to the presently heterogeneous and large-scale communication systems.
	
	To capture the specifics of computer and communication networks, Rusek et al.~\cite{rusek2019unveiling} adapt the MPNN model proposed by Rusek and Cho{\l}da~\cite{rusek2018message} to learn relevant information about the interrelation between network links and network traffic. The  interchange of messages is bidirectional between paths and the links they traverse. Path-level aggregation is a simple summation. Link-level messages are aggregated using RNN to account for sequential dependence (due to packet losses) in the links that form every path. The readout function is obtained from a neural network as in \cite{rusek2018message}. The update function and RNN are modeled using GRU. The model is supervised and  designed for path-level inference purposes (i.e., for predicting delay and jitter metrics). Rusek et al.~\cite{rusek2019unveiling} note that it is straightforward to produce link-level inferences too (for instance, congestion probabilities). A limitation of the model is that it supports only topologies with identical link capacities. The model has been further explored by Suarez et al.~\cite{suarez2019challenging} and extended by Badia et al.~\cite{badia2019towards} to model nodes (their queue size) too.
	
	The authors train their model solely on NSFNet topology with $14$ nodes \cite{nettopo}. The training data comprises various routing schemes and traffic intensities. 
	The test evaluations are performed on the same train topology as well on Geant2 with 24 and GBN with 17 nodes, \cite{knight2011internet}. The  results from this experimental setting suggest good generalization performance, although some decrease in the prediction accuracy of the delay and jitter metrics is observed for unseen and for larger size topologies.

	\subsubsection{Routing policy optimization}

	The network routing optimization problem addressed by Almasan et al. \cite{almasan2019deep} consists of a (central) controller in the control plane of a software defined network that needs to make routing decisions while maximizing the traffic volume routed through the network (an $\mathcal{NP}$-hard problem). The solution proposed by Almasan et al.~\cite{almasan2019deep} integrates MPNN with a deep reinforcement learning. The authors highlight that earlier approaches based on the same basic ideas (a neural network and reinforcement learning) have failed to generalize to unseen topologies. They attribute this to the deep neural network employed (convolutional or fully connected), which does not account for the graph structure of the communication networks.  
	
	The learning agent, which resides in the control plane, has a (global) view of the network and receives traffic requests with different bandwidth requirements. The learning is guided by the optimization objective of maximizing the traffic served by the network. The agent implements the deep Q-learning algorithm. Almasan et al.~\cite{almasan2019deep} employ an MPNN, as is done by Rusek et al.~\cite{rusek2019unveiling}, with the goal to learn and model the relations of links and graphs with routing paths and associated traffic. At each time step, the agent is fed with the state of the network and the new traffic arrival demand. The objective of the MPNN model is to estimate the $q$-value determined by routing the traffic demand considering the current state. Since the number of possible routing solutions to each pair of source and destination results in a high-dimensional space (especially in large-scale real-world networks), the number of  $q$-values that the agent needs to estimate is correspondingly high too \cite{almasan2019deep}. Almasan et al. overcome this problem by limiting the set to $k$ candidate paths. The actions of the agent are introduced into the MPNN, which makes the action representation invariant to permutations of nodes and edges. The goal is the trained MPNN to understand actions over unseen network states and topologies. For training, an $\epsilon$-greedy exploration strategy is applied, which either makes a random decision (with a probability of $\epsilon$) or executes the action with the highest $q$-value.
	
	The  model is trained on the 14-node NSFNet topology \cite{nettopo}. A traffic demand is generated uniformly at each time step by selecting a source-destination pair and traffic demand bandwidth \cite{almasan2019deep}: the agent knows past and present but not future demands. The authors report that their model scales when evaluated on a 24-node Geant2 topology~\cite{nettopo} and outperforms a chosen deep reinforcement baseline. When tested on real-world topologies from the Internet Topology Zoo data set \cite{knight2011internet} with a number of nodes $n=[5,50]$, the model is reported to generalize. Results from experiments, which involve link failures, suggest that the model is robust to such events too.

	\subsection{Scheduling}
	\label{sec:netw-scheduling}
	
	 \begin{table*}[!ht]
		\begin{center}
			\begin{tabular}{|l|l|l|l|l|l|}
				\hline
				\multicolumn{6}{|c|}{\textbf{Scheduling}}\\
				\hline
				\textbf{Problem} &\textbf{Reference} & \textbf{Learning} & \textbf{Approach} & \textbf{Performance summary} & \textbf{Remarks}   \\
				\hline
				Cluster&Mao et al.&reinforcement& GNN + custom &Learns scheduling policy by exploiting &Generalizes to arbitrary job DAG shapes \\
				Scheduling&\cite{Mao2018}&&policy network & workload structure. Outperforms base-& and sizes. Scales to unbounded input arrival \\
				&&&&line heuristics on their performance & sequences. Applicability extends to use as  \\
				&&&&metrics. &CPU or memory scheduler.\\
				\hline
				Wireless&Wang et al.&reinforcement&A3C network&Improves QoE by 30\% and reduces & Centralized system. Optimizes for personal-\\
				edge  &\cite{Wang2020a, Wang2019}&&based& penalty (computation + bandwidth& ized QoE. Generalizes to unseen traffic and  \\
				content &&&& cost) by 35\% compared to baseline & different edge capacities. Train and tested \\
				scheduling &&&&heuristics.&on one network with real data. Re-train \\
				&&&&&required to for different networks.\\
				\hline
				Packet&Zhang et al.& reinforcement&LSTM + DQN& Outperforms baseline methods (round& Train and tested on a two-node network \\
				scheduling &\cite{Zhang2019}&&&robin, MinRTT~\cite{Zhang2019}, BLEST~\cite{Ferlin2016blest}, &with two subflows. Adapts to different  \\
				multi-path &&&&DEMS~\cite{Guo2017dems}). Improves goodput by & network delays and variances in \\
				TCP &&&&10\%-20\% compared to MinRTT.&bandwidth.\\
				\hline
			\end{tabular}
		\end{center}
		\caption{Representative approaches implementing previously addressed ML algorithms for solving scheduling problems within networking.} 
		\label{tab:CoinSched}
	\end{table*}

	Scheduling is the process by which work-elements are assigned to resources to complete a high-level objective.
	The process involves determining the mapping from work-elements to resource instances and the order at which mapped work-elements are processed. 
	An overview of the scheduling approaches described in this subsection is provided in Table \ref{tab:CoinSched}.
	
	\subsubsection{Cluster Job Scheduling}
	
	Data processing clusters provide users with computing power to process their workloads (\emph{jobs}). Each job consists of processing stages (in turn consisting of several tasks) connected by input-output dependencies, which allow jobs to be modeled as direct acyclic graphs (DAGs). 
	The goal of the cluster scheduler is to assign jobs' stages to available resource instances (executors).
	It must be able to scale to thousands of jobs and hundreds of machines, thus deciding among potentially hundreds of configurations per job. This may lead to larger problem sizes compared to conventional reinforcement learning applications like game-playing. 
	Even relaxing the problem to DAGs having homogeneous tasks or an independent set of heterogeneous tasks leads to an NP-hard problem~\cite{Grandl2016}.
	
	Mao et al. \cite{Mao2018} developed a method to learn workload-specific scheduling algorithms used by a cluster scheduler.
	The authors argue that current approaches are inefficient since they rely on the use of simple and generalized heuristics and ignore the rich and easily-available job structure for performing scheduling decisions. Mao et al. \cite{Mao2018} propose Decima, a reinforcement learning scheduling agent that combines a GNN to process job and cluster information, and a policy network responsible for making the scheduling decisions. 
	For realizing this, Decima overcomes three challenges: process state information in a scalable manner, efficiently explore a vast space of scheduling decisions and deal with continuous stochastic job arrivals. Scalability of state information processing is achieved through the use of a GNN responsible for encoding the environment state information through a set of three different types of embedding vectors: per-node embeddings (capture information about a DAG node and its neighbors), per-job embedding (aggregate information across a DAG) and a global embedding (summary of per-job embeddings).
	Embeddings are computed by message-passing operations from leaves of the direct acyclic graphs to its parent nodes. In contrast to conventional GNN architectures \cite{COoverGraphs17}, Decima adds a second non-linear transformation to the conventional node embedding operation. 
	For training, stochastic continuous job arrival is supported by terminating initial training episodes early and gradually increasing the episode length.
	Decima was implemented on a 25 node Spark cluster, whereby the number of executors per node is determined by the Spark master based on Decima's output.
	Experiments show that Decima reduces average job completion time by at least 21\% on average-load and 200\% on high-load case compared to the best performing heuristic (optimized weighted fair scheduling \cite{Mao2018}). Decima was also tested for CPU and memory-scheduling and outperforms prior schemes, such as Graphene \cite{Grandl2016}, by at least 30\%.
	Scalability with respect to both the number of jobs to be processed and variations in jobs workload structure is assured by the use of GNN to generate embedding vectors at different abstraction levels. Scalability is complemented by instantiating a Decima agent for each application generating jobs. In contrast to heuristic methods, the ability of Decima to exploit workload structure ensures adaptability to unseen jobs. Furthermore, its adaptation for CPU and memory-scheduling extend the approach applicability to other areas different than cluster scheduling.
	
	\subsubsection{Edge-Layer Content Scheduling}
	
	Scheduling at the edge of wireless infrastructures involves serving client sessions through different queues at the edge-layer (last-hop link). Here, several numbers of heterogeneous applications, even within a client, share the same resources under different channel characteristics for operation. The problem complexity scales with the number of clients, the number of queues per device, and the number of edge devices in the wireless network.
	
	Wang et al. \cite{Wang2020a, Wang2019} optimize towards personalized Quality of Experience (QoE) in the use-case of crowdsourced livecast (crowdcast) edge content distribution. Crowdcast is characterized by highly diverse viewer side content watching preferences and environments. While some users care more about channel switching latency, as they browse many channels frequently, fidelity users regularly watch one channel and are insensitive to it. These aspects make it especially challenging to accommodate heterogeneous and personalized QoE.
	The authors argue that existing edge resource allocation approaches use predefined rules or model-based heuristics to make scheduling decisions.
	Integrating users' personalized QoE-demands results in a NP-complete problem~\cite{Wang2020a}.
	Wang et al. \cite{Wang2020a} developed DeepCast, a reinforcement learning edge-assisted framework combined with a deep neural network for performing scheduling policy based on real-time network information to accommodate personalized QoE. The model learns suitable strategies for scheduling and transcoding using an A3C network model \cite{pmlr-v48-mniha16}. In DeepCast, the edge-server receives the high-bitrate content from the content distribution network, which, if needed, is transcoded (downsampled) at the edge layer before delivering it to the end-user.
	For evaluation purposes, Wang et al. \cite{Wang2020a} perform trace-driven experiments by integrating three real-world datasets comprising viewing, location, and bandwidth information, respectively. DeepCast aims at minimizing a penalty function that considers QoE, computation cost (from transcoding operations at the edge), and system bandwidth cost. Served users are assumed to connect or disconnect dynamically (online scenario).
	The results were compared with traditional content distribution network architecture and other learning-based methods that consider either part of the QoE metrics or the system cost. In the cases when all the QoE metrics are included for training, DeepCast outperforms learning-based approaches (DQN considering only a subset of QoE metrics) by at least 30\%. 
	For the optimality analysis, the authors first compute the optimal analytical solution through an offline scenario (served users come in periodical batches, not dynamically) and compare it with results from both DeepCast and an online greedy heuristic that selects best server and bitrate for each coming user individually. Compared to the baseline heuristic~\cite{Wang2014jointonline}, DeepCast reduces the gap to the offline-optimum by 35.3\% on average. For generalization analysis, the authors focus on generalization concerning traffic patterns, not network size. Wang et al. \cite{Wang2020a} consider a new scenario in which synthetic data is used to train DeepCast, and real trace data is used for testing purposes. Compared to the online heuristic~\cite{Wang2014jointonline}, DeepCast reduces the penalty by 5\% to 10\% in this generalization scenario. 
	Although the authors analyze the impact of the edge capacity, the ratios of the penalty weights in the penalty function, and the learning model (A3C network vs. DQN) in DeepCast performance, the impact of the network size is not addressed. Network size has a direct influence in the state space and action space that needs to be processed. The authors suggest that DeepCast can be deployed at each regional server requiring a small overhead in training update.
	
	\subsubsection{Packet Scheduling}
	
	A further instance of scheduling is present when assigning packets from the same data source to different paths connecting a source-destination node pair within the network, as is the case of multi-path TCP when the routes are known a priori.
	The scheduler then determines the number of packets distributed among the possible subflows for each of the sessions. The problem scales with the number of sessions and their respective subpaths.
	It is even more challenging when considering network heterogeneity due to possible substantial differences across multiple subpaths (in terms of delay and capacity), which may require a long buffer queue at the receiver to reorganize received packets. 
	
	Zhang et al. \cite{Zhang2019} introduce a learning-based scheduler that adapts to dynamic heterogeneous network environments. They argue that current approaches are far from optimal, since they lack self-adaptivity and optimize only towards a specific performance metric, e.g., minimizing head-of-line blocking, or utilize simple heuristics, e.g., prioritizing subflows with shortest RTT. The authors propose a reinforcement learning based Scheduler (ReLeS) that learns the control policy for packet scheduling through a neural network. 
	ReLeS uses stacked LSTM networks to extract features over a look-back of $k$ past states. These are then fed to a policy network that outputs the data split ratios among the different subflows. ReLeS is trained asynchronously and iteratively offline after observing the reward of scheduling decisions performed online. The reward function is designed to address network heterogeneity challenges and aims at maximizing aggregated bandwidth among subflows while minimizing average packet delay and the number of lost packets during training.
	ReLeS evaluation is performed using a single source-destination pair with two subflows. Results are compared with existing schedulers such as MinRTT (default scheduler for multi-path TCP), Round Robin, BLEST~\cite{Ferlin2016blest}, and DEMS~\cite{Guo2017dems}. ReLeS outperforms these schedulers in most cases, with a 10\%-20\% increase of goodput compared to MinRTT. 
	Since the network output represents the split rations between the subflows of source-destination pair, it suggests that either one ReLeS is trained for each source-destination pair or that all source-destination pair must have the same number of subflows, thus limiting scalability with respect to the number of subflows.
	Moreover, the authors demonstrated ReLeS's adaptability to changing network delays and variances of bandwidth while maintaining good performance compared to baseline methods.

	\subsection{Resource Allocation}
	
	\label{sec:netw-resourceallocation}
	
	Resource allocation is the process by which network resources (e.g., memory, CPU, or bandwidth) are provisioned to fulfill applications' requests or requirements. It can be understood as a previous step to the scheduling action. 
	An overview of the resource allocation approaches described in this subsection is provided in Table \ref{tab:CoinResAlloc}.

	 \begin{table*}[!ht]
		\begin{center}
			\begin{tabular}{|l|l|l|l|l|l|}
				\hline
				\multicolumn{6}{|c|}{\textbf{Resource allocation}}\\
				\hline
				\textbf{Problem} &\textbf{Reference} & \textbf{Learning} & \textbf{Approach} & \textbf{Performance summary} & \textbf{Remarks}   \\
				\hline
				VNF&Mijumbi et al.&supervised&GNN based&Predicts VNF component resource & Trained on a five-node graph. Scalability to\\
				deployment&\cite{Mijumbi2017, Mijumbi2016}&&&requirements. Able to improve system & larger topologies may be limited by \\
				&&&&performance on evaluated use-case.&memory constraints. Generalizes to unseen  \\
				&&&&&data (90\% test accuracy).\\
				\hline
				Network&Liu et al.&reinforcement&DQN based& 2x to 4x orders of magnitude improve- & Decentralized: multiple agents coordinated \\
				slicing in&\cite{Liu2020}&&&ment compared to traffic aware resource &by one. Scales to different network sizes \\
				edge&&&&orchestration baseline&and number of slices. \\
				computing&&&&&\\
				\hline
				Fog&Mseddi et al.&reinforcement&DQN based& Delivers close to optimal solutions while & Increasing load results in longer training. \\
				computing&\cite{Mseddi2019}&&&outperforming nearest and random fog &Better generalization to different mobility \\
				&&&&node association \cite{Mseddi2019}.& scenarios (compared to baseline methods)\\
				\hline
			\end{tabular}
		\end{center}
		\caption{Representative approaches implementing previously addressed ML algorithms for solving resource allocation problems within networking.} 
		\label{tab:CoinResAlloc}
	\end{table*}

	\subsubsection{Virtual Network Function Orchestration}
	
	Network Function Virtualization (NFV) allows for separating the provisioning of network functions, e.g., firewall, from traditional network appliances (hardware) through virtualization towards reducing operational and capital expenses.  
	In practical terms, a VNF is typically instantiated on one or several virtual machines, one for each VNF component, that may be horizontally or vertically scaled. Sequentially coupling several VNFs is referred to as Service Function Chaining (SFC). In SFC, resources from a VNF component may depend on its adjacent VNF components, and its sequential nature allows them to be modeled as a forward graph.
	
	Mijumbi et al. \cite{Mijumbi2016, Mijumbi2017} address the issue of autonomously, efficiently, and dynamically allocating physical resources to VNFs. The authors argue that several approaches have been proposed for VNF placement, but not for the dynamic allocation of their resources. As the load on a single VNF can vary over time, there is a need for dynamic and automated scaling of resources, while considering the topological information on how the VNFs are connected within the SFC.
	Mijumbi et al. \cite{Mijumbi2017, Mijumbi2016} propose an automatic and topology-aware approach to manage VNF resources within SFC based on GNN, a VNF forward graph. The use of GNN forms a connectionist approach, allowing each VNF component to consider resource utilization from neighboring VNF components to manage resources. Each VNF component is modeled as a node in a GNN composed of two parametric functions, each implemented by a neural network. The first models the dependence of a VNF component resource requirements from that of its neighbors, whereas the second forecasts resource requirements of the VNF component. 
	The goal of each of these neural network pairs is to learn, in a supervised manner, the trend of resource requirements of the VNF component that they model \cite{Mijumbi2017, Mijumbi2016}. Once trained, both neural networks of each VNF component predict future resource requirements, allowing the system to perform vertical or horizontal scaling of virtual machines dynamically.
	The authors evaluated their approach on a system implementing a single VNF forward graph consisting of six nodes, each representing a VNF component hosted in one or several virtual machines. 
	They argue that while they performed their experiments on only one VNF forward graph, their approach is scalable to any number of VNFs as long as an SFC's topology can be created. Only horizontal scaling of virtual machines was considered based on CPU utilization predictions, where the last 20 observations on a VNF component were used to predict the resource requirements for 20 time-steps in the future.
	The system is able to achieve a 90\% prediction accuracy on the test set after training using the backpropagation through time algorithm.
	The results were compared to the static scenario (i.e., no scaling at all) and to a manually programmed deployment (i.e., scaling based on fixed thresholds). Their system reduces the processing delay by 27\% and calls drop rate by 29\% in the observed use-case. The authors recognize the further improvement in terms of test accuracy to better generalize to unseen training examples and conjecture that including the effect of traffic arrivals in the prediction might improve the accuracy and adaptability of the system.
	Furthermore, Mijumbi et al. \cite{Mijumbi2017} suggest studying the applicability of their solution on larger SFCs, since there might be memory limitations for representing VNF component states in those cases.
	
	\subsubsection{Network Slicing}
	
	Resource allocation for network slicing allows for multiple logical networks (i.e., slices), each with its performance requirements, to run on top of a shared physical network. It is of particular interest in 5G networks, where heterogeneous services with highly diverse performance requirements (e.g., delay, bandwidth, and reliability) must be accommodated in a scalable and cost-efficient manner.
	
	Liu et al. \cite{Liu2020} present a decentralized deep reinforcement learning approach for slicing wireless edge computing networks, EdgeSlice. According to the authors, existing works on multi-resource allocation are not efficient since it is generally assumed that resources are allocated according to certain ratios, e.g., between radio spectrum and computing resources. 
	EdgeSlice consists of multiple distributed agents that learn the orchestration policy under the coordination of a centralized performance coordinator. Each agent is placed on a set of network infrastructures, consisting of base stations and edge servers, and operates on a second-based timescale. The central performance coordinator ensures the fulfillment of both system capacity and service level agreement, operating on a much larger timescale.
	Decentralized agents in EdgeSlice are trained using deep deterministic policy gradient ~\cite{Lillicrap2016ddpg}, which trains an actor and a critic network, both implemented by a DQN.
	Liu et al. prototyped EdgeSlice on a system composed of a RAN with two eNodeBs (operating at different frequency bands), six OpenFlow switches as transport network, a core network (one desktop computer), and two edge servers. Orchestration agents and performance coordinator were implemented in the core network. Both actor and critic networks of the orchestration agent are composed of a two-layer fully-connected neural network, trained offline using a simulated network environment.
	The use-case for evaluation is that of a mobile application offloading a computation task (object detection video analysis) to the edge server, where each of the two network slices uses different frame resolution. EdgeSlice is compared to the traffic-aware resource orchestration algorithm and EdgeSlice-NT (a simplified version that does not use queue length of network slices in the state description of the model). EdgeSlice obtains a 3.69x and 2.74x improvement on system performance (slice dependent metrics such as latency, throughput, queue status) over traffic-aware resource orchestration and Edgeslice-NT, respectively. Furthermore, the authors study the degree of scalability in a simulated environment by increasing both the number of network infrastructures (regions composed of edge servers and base stations) and the number of network slices using a telecommunication network trace dataset. EdgeSlice is able to generalize to the network size when increasing the number of network infrastructures without notably sacrificing performance \cite{Liu2020}, but not when increasing the number of network slices. However, the system still outperforms traffic-aware resource orchestration on both scenarios. Moreover, the results, both with the prototyped and the simulated environment, suggest that the system is able to adapt to different applications and traffic characteristics.
	
	\subsubsection{Computing Resources in Wireless Networks}
	
	Fog layer devices (e.g., vehicles, access points, routers) can communicate and cooperate to perform storage and processing tasks. 
	The behavior of these fog nodes is highly dynamic since nodes can arbitrarily go online-offline, their resource availability changes drastically over time, and additional mobility patterns of some devices (e.g., vehicles, smartphones) need to be considered.
	
	Mseddi et al. \cite{Mseddi2019} propose a deep reinforcement learning based online resource allocation approach for dynamic fog computing.
	According to the authors, current learning approaches for online resource allocation are either developed for cloud computing or ignore mobility and workload variations of the fog nodes.
	The authors provide a mathematical problem description, which can be proven to be NP-hard, and employ a DQN to solve it, integrating mobility and computing resource availability of fog devices. The state of the proposed framework includes the resource requirements for a given user request, as well as the current state of all fog nodes. The action returned by the DQN corresponds to the fog node index, indicating where to offload the computation.
	The model aims at maximizing the number of satisfied user requests subject to QoS requirements (predefined delay threshold) and limitations of fog nodes.
	Mseddi et al. \cite{Mseddi2019} evaluate their model on real-world data over a broad mobility spectrum, including stationary (roadside units), slow-moving (pedestrians walking), and fast-moving (cars and passengers in buses and trains) fog nodes.
	The evaluation results are compared to two baseline methods~\cite{Mseddi2019} (nearest fog node allocation and random fog node allocation) and to the optimal solution obtained by solving with CPLEX. In the static scenario, all approaches lie close to the optimal solution, but the introduction of dynamic scenarios degrades system performance. However, the approach outperforms baseline methods, having a 10\% degradation to the optimal solution compared to at least 25\% degradation of baseline methods.
	The proposed approach faces some limitations in terms of scalability. First, both state and action of the proposed framework are dependent on the number of fog nodes in the network. Moreover, the training iterations required for convergence increases significantly with the traffic load, which the authors attribute to a higher dimensional system, but no concrete metric is provided as to which extent. The approach main advantage is its adaptation to different mobility scenarios, where it reached the closest to optimal solution compared to the two baseline methods.

	\subsection{Discussion}
	\label{sec:netw-discussion}
	Analyzing the surveyed research contributions, we observe that machine learning approaches at large are capable of outperforming baseline approaches in terms of optimal performance. Important design criteria (Section~\ref{sec:MLforCOP}) related to scalability, adaptability, and ability to generalize over unseen input are in this set of surveyed work evaluated to a varying degree, with emphasis on scalability and generalization abilities. 
	
	We also see that in our list of works, GNN is the dominating approach followed by deep reinforcement learning (Table~\ref{tab:NetworkingOverview}). Although the list is non-exhaustive, we believe this is indicatory of how today's machine learning approaches for networks are primarily designed. Scalability is an essential aspect of networking, and the application of GNN is a key enabler to achieve that - moreover, networks are ultimately discrete event systems, well matching the properties of reinforcement learning approaches. Further, we observe a tendency that reinforcement learning algorithms appear to be more widely applied than supervised learning. One explanatory factor from the perspective of machine learning methods design is that adaptability and generalizability (and altogether robustness to perturbation and unseen data) can be inherently incorporated in reinforcement learning algorithms, compared to supervised learning. Further, in comparison with reinforcement learning approaches, supervised learning generally requires labeled input that also adheres to the problem instance and puts stronger requirements on the availability of enough data to generalize over, which is often not the case in a practical setting. Although networking infrastructures and services generate large volumes of data, preprocessing and labeling takes substantial efforts - hence, supervised learning methods are generally less applicable to many resource management problems and practical settings.
	
	From the perspective of algorithm execution times, it is less clear whether machine learning algorithms are generally faster than baseline heuristics or solvers (such as CPLEX). However, although the execution time relates to the problem size and data volume (e.g.,~\cite{Mseddi2019}), most of the applied machine learning approaches can be run by parallelized algorithm execution, if required for the practical conditions at hand.
	
	Finally, the set of illustrative machine learning approaches for resource management in networks is limited to only a few methods, in comparison to what has been shown available in Section~\ref{sec:MLforCOP}. With awareness of applicable methods evaluated in other domains, we see that advances in solving many different resource management problems could benefit from broadening the set of applied machine learning approaches for combinatorial optimization that have been evaluated with promising results. Moreover, systematic evaluation of several important design aspects for machine learning, primarily concerning scalability, adaptability, and generalization, is essential to ensure practical applicability of the worked-out solutions.

	\section{Summary and outlook}

	Combinatorial optimization problems on graphs are paramount for numerous and diverse practical domains. 
	Therefore, such problems need to be solved efficiently in terms of execution time and optimality of the results. 
	Furthermore, as they arise recurrently, it is desirable to automate the design of such algorithms. 
	We surveyed the potential of machine learning approaches for achieving these goals. 
	Reinforcement learning and supervised learning are competitive to each other, considering performance results on combinatorial optimization tasks. Learning structures that emerge as contemporary solutions to such demanding problems are attention mechanisms and graph neural networks. Promising results in terms of optimality, run time, and generalization have already been obtained with some of the surveyed machine learning approaches. 
	Nevertheless, for the learning algorithms to become a standard tool for overcoming combinatorial optimization challenges, future research must comprehensively address several relevant aspects: scalability, adaptability, generalization, run time, and automation (we discussed them at length). For practical purposes, distributed machine learning is also essential for the successful implementation of machine learning for combinatorial optimization in applied domains such as the networking. These questions are high on the machine learning research agenda. However, considering  the progress made in the past few years in solving combinatorial optimization problems by learning algorithms, we could expect practically applicable machine learning models to be innovated within this active research field in the foreseen future.

	\section*{Acknowledgment} The authors are grateful to the IEEE Access Associate Editor Dr. Junaid Shuja, who coordinated the review process,  and to the anonymous reviewers for their expeditious and helpful review comments and suggestions received in preparation of the final version of the manuscript.

	\appendix
	
	\section{Combinatorial optimization problems on graphs}
	\label{sec:Appendix}
	
	\textbf{TSP.}
	The Travelling Salesman Problem (TSP) is defined by a set of cities and the goal is to find the shortest path (the route with minimum cost) that visits each city only once and returns to the starting point (initial city). Formally, TSP is defined on an undirected edge-weight graph $G = \langle V, E\rangle$, where $V$is the set of nodes, $E$ is the set of edges connecting the nodes in $V$ to be visited and $w(e)$ is the weight (cost) of edge $e$. The set of feasible solutions $F$ is composed by all edge subsets that determine a Hamiltonian cycle\footnote{A Hamiltonian cycle is a closed tour on a graph, which involves visiting every node exactly once.}, and the objective value $c(f)$ to be minimized is the sum of the weights  $w(e)$ of all edges $e \in E$ in a solution $f \in F$: $c(f) = \sum_{e \in E}w(e)$.
	The TSP problem arises in several practical domains, most notably transportation, communication networks, scheduling, planning, and as noted by Vinyals et al. \cite{PointerNets15} in microchip design or DNA sequencing too. TSP can be solved by dynamic programming in $\mathcal{O}(|V|^2 2^{|V|})$ time. The time complexity of TSP and its variations is discussed in detail by Gutin and Punnen in \cite{gutin2006traveling} within the scope of exponential neighborhoods and domination analysis.
	
	\textbf{VRP.} 
	The Vehicular Routing Problem (VRP) is a common extension to the TSP. In this setting, the aim is to find the optimal sequence of nodes (cities) such that the total tour cost (length of the route, time, or number of vehicles among other possible criteria) is minimized while all other constraints such as capacity remain fulfilled. In its basic form, each node is visited only once by only one vehicle with a finite capacity. A complete graph is given $G = \langle V, E\rangle$, where one of the nodes $v \in V$ is the depot, and the rest are cities to be visited. The weight of an edge $w(e)$ reflects its cost. A demand is associated with each node and the capacity of a vehicle is at least as large as a node demand. The goal (as in the TSP case) is to minimise the total cost:  $c(f) = \sum_{e \in E}w(e)$ subject to the aforementioned constraints.
	Vidal et al. in \cite{vidal2013heuristics}  survey VRP (including meta-heuristics) and evaluate the computational complexity of some specific VRP solutions.
	
	\textbf{MaxCut.}
	Maximum Cut (MaxCut): Given an undirected graph $G = \langle V, E\rangle$ and non-negative weights $w_{ij} = w_{ji} \geq 0$ of the edges $(i,j) \in E$, find the subset of vertices $S \in V$ that maximizes the weight of the edges in the cut $(S, \bar{S})$; that is, the weight of the edges that connect a node in $S$ with a node in $\bar{S}$: $\max w(S, \bar{S}) = \sum_{i \in S, j \notin S} w_{ij}$. 
	The problem, as noted by Barret et al.~\cite{barrett2019exploratory}, can appear in different applied domains such as computational biophysics (protein folding, which has implications to medicine), finance (investment portfolio optimization) and physics.
	The time-complexity of the simple MaxCut problem  has been evaluated by Bodlaender and Jansen in \cite{bodlaender2000complexity} for various special classed of graphs. The authors prove that simple MaxCut problem can be solved in $\mathcal{O}(|V|)$ time on graphs with constant bounded treewidth and that there exists an $\mathcal{O}(|V|^2)$ algorithm on cographs (see   \cite{bodlaender2000complexity} for a mathematical definition of a cograph).
	
	\textbf{MVC.} 
	Minimum Vertex Cover (MVC): Given an undirected graph $G = \langle V, E\rangle$, find the smallest subset of nodes $V_c \subseteq V$
	such that each edge in the graph $e \in E$ is incident to at least one node in the selected set $v \in V_c$. In other words, every edge has at least one endpoint in the vertex cover $V_c$. Such a set covers all edges $e \in E$ of $G$.
	A minimal vertex cover corresponds to the complement of maximal independent vertex set (defined below).
	MVC arises in the field of biochemistry and biology \cite{chen2006improved}, for instance.
	For an idea of the worst time complexity of the Vertex Cover (VC) problem and some recent improvements on the upper bound, see the work of Chen et al.~\cite{chen2006improved}. Specifically, Chen et al. propose an $\mathcal{O}(1.2738^k + kn)$-time polynomial-space parametrization algorithm for VC.  
	
	\textbf{MC.}
	Maximal Clique (MC): Given an undirected graph\footnote{An \textit{undirected} graph has bidirectional edges in contrast to directed graphs, where edges can be traversed only in one direction.} $G = \langle V, E\rangle$, find the largest subset of vertices (nodes) $C \subseteq V$ that form a clique. A clique of $G$ is a complete\footnote{In a \textit{complete} graph, each pair of nodes is connected with a unique edge.} subgraph of $G$. In other words, MC is the largest subset of nodes in which every node is directly connected to every other node in the subset. An MC cannot be extended by adding one more adjacent node, that is, it is not a subset of a larger clique. 
	The MC problem can be encountered in many diverse applications such as  in clustering and in bioinformatics (in problems related to molecular structures). 
	There are evolutionary algorithms that implement heuristics for MC such as those proposed by Guo et al.~\cite{guo2019meamcp}.
	
	\textbf{MIS.}
	Maximal Independent Set (MIS): Given an undirected graph $G = \langle V, E\rangle$, find the largest subset of vertices $U \subseteq V$ in which no two are connected by an edge; that is, for any vertex pair ($u_i, u_j$)  in $U$,  $(u_i, u_j) \notin E$.
	MIS finds application in classification, coding theory, geometry, and VLSI design among other~\cite{das2012heuristics}.
	Exact algorithms for the MIS problem have been developed by Xiao and Nagamochi \cite{xiao2017exact}, for instance, who propose a $1.1736^{|V|} |V|^{\mathcal{O}(1)}$-time algorithm for MIS in a graph with  $|V|$ nodes and degree bounded by~5. A recent heuristic algorithm for solving MIS with an absolute accuracy estimate and time-complexity $\mathcal{O}(|V|^2|E|)$ has been proposed by Gainanov et al. \cite{gainanov2018heuristic}. Das et al.~\cite{das2012heuristics} discuss various evolutionary approached for solving MIS (without evaluating their complexity).

	\textbf{MCP.}
	Maximum Coverage Problem (MCP): The unweighted MCP is defined as follows. Given a set of sets $\mathcal{U} = \cup_{n=1}^{m}S_n$ and a number $k$, select at most $k$ sets $S_{i_1}, S_{i_2}, ..., S_{i_k}$ from $\mathcal{U}$ such that the number of covered elements, i.e., the cardinality of the union of the sets $|\cup_{j=1}^{k} S_{i_j}|$, is maximized. Observe that the sets in $\mathcal{U}$ may have some elements in common. A budget version of the MCP reads as follows. Given a bipartite graph over two sets of nodes $ V = V_1 \cup V_2$ and a budget $b$, select a subset of nodes $S \in V_1$, $|S|=b$  such that a maximum number of nodes from $V_2$ have at least one neighbor from $S$.
	Examples of heuristic algorithms devised for MCP are discussed by Chandu~\cite{chandu2015big} and of meta-heuristics by Bilal et al.~\cite{bilal2013new}, who evaluate the run time of the algorithms under various experimental settings.
	
	\textbf{GC.} Graph Coloring (GC): Given an undirected graph, find the minimum number of colors with which the nodes can be colored such that no edge connects two nodes with the same color. The problem arises in planning, scheduling, and allocation, to name a few. The graph coloring problem can be mapped to the \textit{scheduling problem} by representing the objects to be scheduled by nodes. An edge connects every two nodes that cannot share the same resources. Different algorithms, including meta-heuristics such as hill climbing, simulated annealing, tabu search, and genetic algorithms, are reviewed and evaluated by Wang and Xu~\cite{wang2013metaheuristics}. Fromin et al.~\cite{fomin2007improved} develop an exact algorithm for the decision version of GC. The algorithm has $\mathcal{O}(1.7272^{|V|})$ run time on the number of nodes $|V|$ for deciding if a graph is 4-colorable. 
	One primary result achieved by Shimizu and Mori is a polynomial-space  quantum algorithm for the graph 20-coloring problem with run time $\mathcal{O}(1.9575^{|V|})$.
	
	\textbf{SAT.}
	Satisfiability Problem (SAT): Consider a Boolean (propositional) logic expression consisting of Boolean variables, parentheses,
	and the operators AND (conjunction), OR (disjunction), and NOT (negation). 
	A literal is (a negation of) a Boolean variable. A clause is a disjunction of literals. A Boolean expression is a finite conjunction of clauses. 
	The SAT problem consists of finding a Boolean assignment to all variables such that the given expression is true, or that no such assignment exists.
	Satisfiability solvers are used as a general-purpose tool in various domains such as combinational equivalence checking, model checking,  automatic test-pattern generation, planning and genetics, scheduling, and combinatorics \cite{marques2008practical}.
	An evolutionary (meta-heuristic) algorithm is proposed by Bougachi et al. in \cite{1460462} and its run time extensively evaluated.

	\textbf{KP.}  The Knapsack Problem consists of a knapsack with a finite capacity $W$ and a set $O = \{o_1, o_2, ..., o_n\}$ of $n$ objects, each of which is associated with a weight $w_j \leq W$ and a profit $p_j$. The capacity $W$, weight $w_j$ and profit $p_j$, without loss of generality, are assumed to be non-negative integers $W, w_j, p_j \in Z^{+}, j=1,...,n$ and $\sum_{j=1}^{n} w_j > W$. The set of feasible solutions $F$ is formed by all object subsets $F_i$ such that the sum of the weights of the elements in $F_i$ does not exceed $W$. The objective function value $c(f)$ to be maximized is the sum of the profits of all objects in a solution $F_i \in F$: $c(f) = \sum_{j\in F} p_j$. 
	The knapsack problem is a \textit{resource allocation with constraints} problem that arises within various domains. A general (financial) formulation is: given a budget select the (maximum number of) items to purchase. In wireless networks, for instance, the budget is the set of available radio resources, and the `items' are the data packets that need to be served (sent) with the objective to maximize the number of packets sent. Other areas where KP arises are cryptography and computer security. KP can be solved by dynamic programming in $\mathcal{O}(nW)$ time. Ezugwu et al. \cite{8678633} evaluate meta-heuristic algorithms for the 0--1 KP in terms of the quality of the obtained solutions and computational time. The authors conclude that branch and bound as well as dynamic programming are superior to genetic algorithms, simulated annealing and greedy search with dynamic programming being slightly more optimal yet slightly slower than branch and bound.

	\bibliographystyle{IEEEtran}
	\bibliography{refs}

	\begin{IEEEbiography}[{\includegraphics[width=1in,height=1.25in,clip,keepaspectratio]{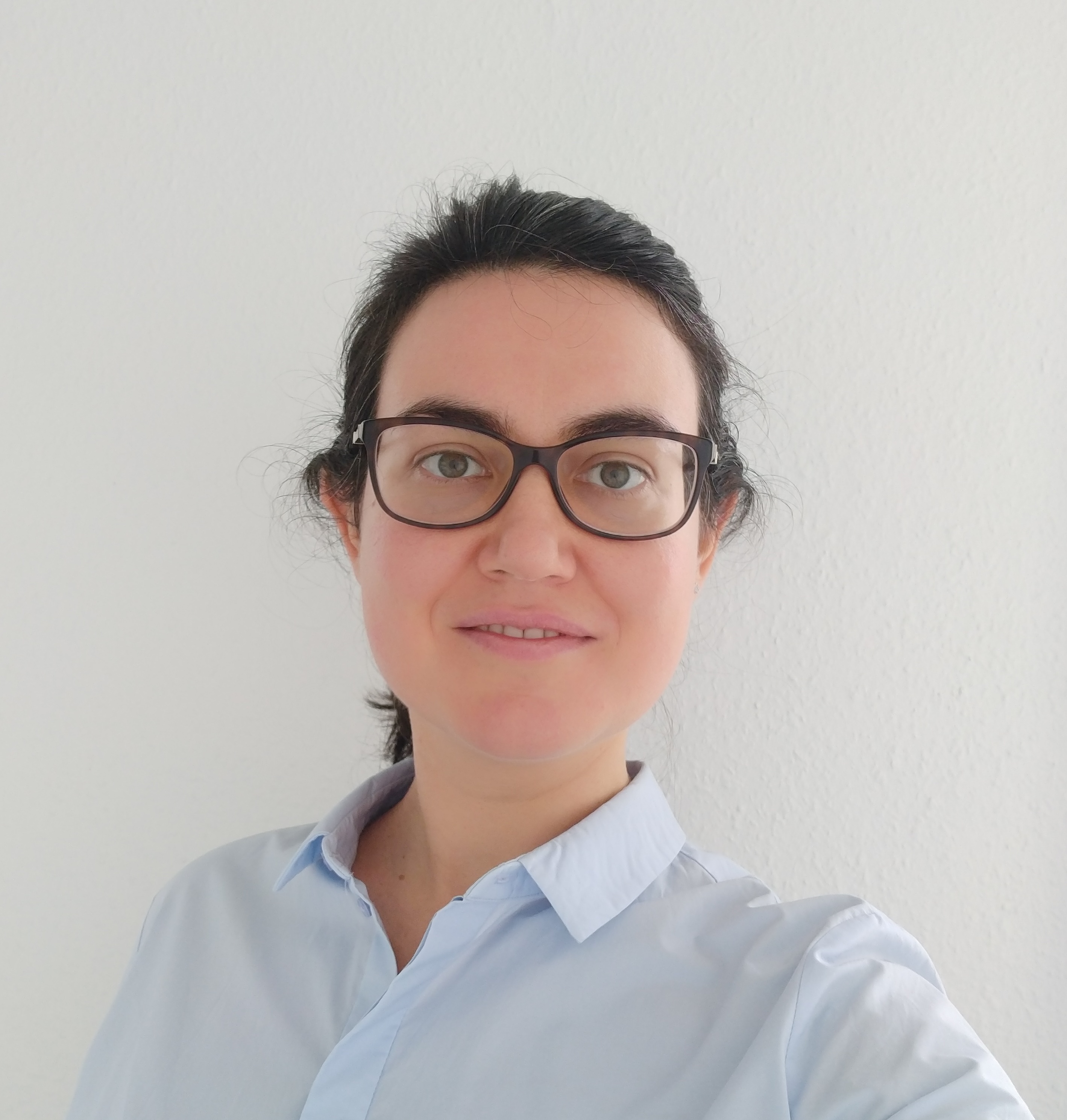}}]{Natalia Vesselinova} accomplished her M.Sc. in telecommunications engineering at Technical University of Sofia, and the Nordic Five Tech M.Sc. in applied and engineering mathematics at Chalmers University of Technology and Aalto University. She gained her Ph.D. in telecommunications at Technical University of Catalonia.
		
	She is a Senior Research Scientist at the Research Institutes of Sweden (RISE), where she contributes to the research endeavors of the Network Intelligence group. At present, she is focused on solving problems from the networking engineering practice with probability, statistics, machine learning and deep learning theory and tools. 
	\end{IEEEbiography}
	
	\begin{IEEEbiography}[{\includegraphics[width=1in,height=1.25in,clip,keepaspectratio]{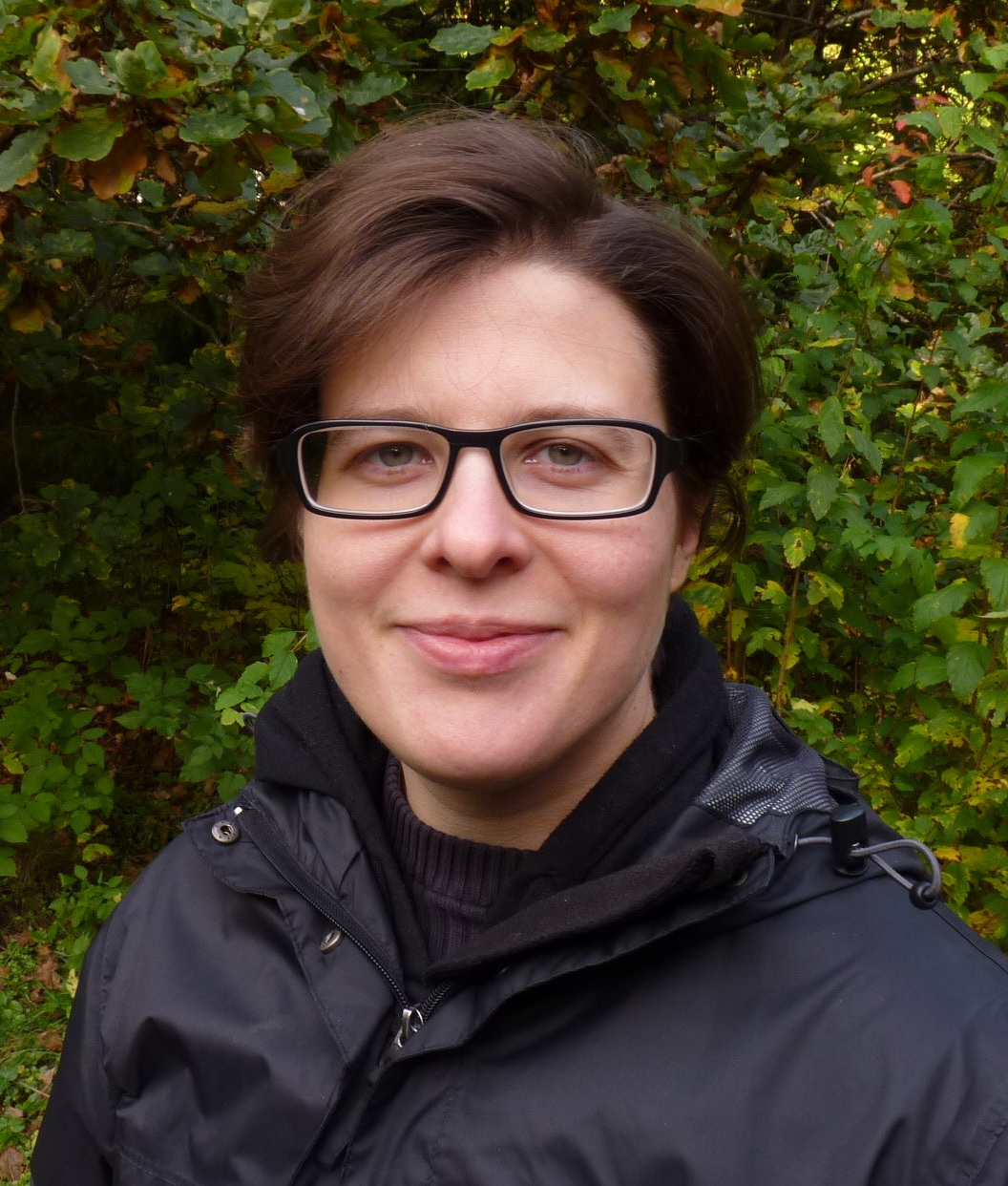}}]{Rebecca Steinert}  holds a B. Sc. in real-time systems since 2002 and received her M.Sc. degree in computer science with emphasis on autonomous systems and machine learning in 2008. In 2014, she finished her PhD in probabilistic fault management and performance monitoring in networked systems at KTH, Royal Institute of Technology, Stockholm. 
		
	She joined RISE Research Institutes of Sweden (formerly SICS Swedish Institute of Computer Science) in 2006. She is currently driving research within applied machine learning for intelligent autonomous networked systems, and is the leader of the Network Intelligence research group since 2015. Her research interests currently include probabilistic modeling, deep learning and combinatorial optimization applied in programmable networks and distributed systems.
	\end{IEEEbiography}

	\begin{IEEEbiography}[{\includegraphics[width=1in,height=1.25in,clip,keepaspectratio]{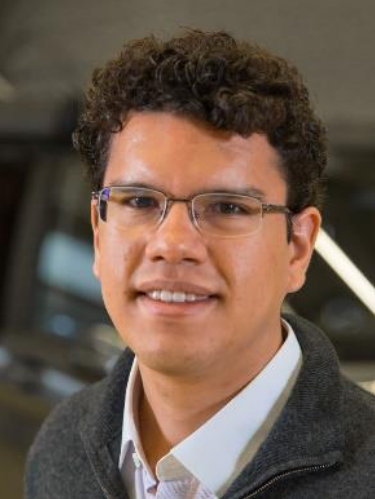}}]
		{Daniel F. Perez-Ramirez}, originally from Colombia, holds a B.Sc. in mechanical engineering and a M.Sc. in Robotics, Cognition, Intelligence with emphasis on cognitive systems and machine learning from the Technical University of Munich (TUM), Germany. He joined RISE Research Institutes of Sweden AB in 2019 at the Network Intelligence research group in Stockholm. Previous to RISE, he worked on applied machine learning for the automotive industry, knowledge representation and sensor integration for robotics applications, and agile methods for mechatronic product development.
		His current research work and interest focuses on applying machine learning and probabilistic models to develop intelligent systems for networking applications.
	\end{IEEEbiography}
	
	\begin{IEEEbiography}[{\includegraphics[width=1in,height=1.25in,clip,keepaspectratio]{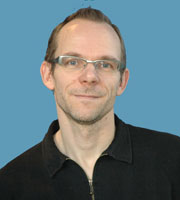}}]{Magnus Boman} 
		is a professor in Intelligent Software Services at the School of Electrical Engineering and Computer Science, 
		at KTH in Stockholm since 2003. He has been putting AI-methods to practical use since the late 1980s, 
		and has a special interest in learning machines that slowly but steadily become more useful over time and over task. 
		In the last ten years, he has focused on health, working in close cooperations with clinicians in epidemiology, 
		cognitive decline, emotion research, and mental health. Boman is an Associate Editor of Eurosurveillance. 
		In his spare time he does reservoir computing and liquid state machine research to support energy-efficient fog computing.
	\end{IEEEbiography}
	
	\EOD
	
\end{document}